\documentclass[preprint,review,12pt]{elsarticle}

\usepackage[style=base]{caption}
\usepackage{pgf}
\usepackage{lscape}
\usepackage{graphicx}
\usepackage{amsmath} 
\usepackage{setspace}
\usepackage{soul}
\usepackage{algpseudocode}
\usepackage{amsmath,amssymb,amsfonts}
\usepackage{mathtools}
\usepackage{times}
\usepackage{fancyhdr}
\usepackage[ruled,vlined]{algorithm2e}
\usepackage{makecell}
\usepackage{tikz}
\usepackage{varwidth}
\usetikzlibrary{arrows.meta}
\usetikzlibrary{calc}
\usepackage[keeplastbox]{flushend}
\usepackage{url}
\usepackage{booktabs, tabularx}
\usepackage[utf8]{inputenc}
\usepackage{xcolor}
\usepackage{hyperref} 
\usepackage{float}
\usepackage[margin=2.6cm]{geometry}

\usepackage{pgfplots}
\pgfplotsset{width=10cm,compat=1.10}
\usepgfplotslibrary{fillbetween}
\include{pythonlisting}

\usepackage{subcaption}[style=base]
\usepackage{soul}

\journal{Applied Soft Computing}

\begin{document}

\begin{frontmatter}


\title{On the performance of deep learning for numerical optimization :an application to protein structure prediction}



\author{Hojjat Rakhshani, Lhassane Idoumghar, Soheila Ghambari, Julien Lepagnot,  Mathieu Br\'evilliers}

\address{Universit\'e de Haute-Alsace, IRIMAS UR 7499, F-68100 Mulhouse, France -- firstname.lastname@uha.fr}

\begin{abstract}
Deep neural networks have recently drawn considerable attention to build and evaluate artificial learning models for perceptual tasks. Here, we present a study on  the performance of the deep learning models to deal with global optimization problems. The proposed approach adopts the idea of the neural architecture search (NAS) to generate
efficient neural networks for solving the problem at hand. The space of network architectures is represented using a directed  acyclic graph and the goal is to find the best architecture to optimize the objective function for a new, previously unknown task. Different from proposing very large networks with GPU computational burden and long training time, we focus on searching for lightweight implementations to find the best architecture. The performance of NAS is first analyzed through empirical experiments on CEC 2017 benchmark suite. Thereafter, it is applied to a set of protein structure prediction (PSP) problems. The experiments reveal that the generated learning models can achieve competitive results when compared to hand-designed algorithms; given enough computational budget.
\end{abstract}

\begin{keyword}
Neural architecture search \sep optimization \sep deep learning \sep convolutional neural network


\end{keyword}

\end{frontmatter}


\section{Introduction}
Optimization algorithms have witnessed prevailing success in different application areas~\cite{8627945,8628260,rakhshani2017snap,RAKHSHANI2019100493}. Formally, they help to find a parameter vector $x^{*}$ in order to minimize an objective function $f(x): \mathbb{R}^{D} \rightarrow  \mathbb{R} $, i.e. $f(x^{*}) \leq f(x)$ for all $x \in {\mathbb{R}}^{D} $, where $D$ denotes the dimensionality of the problem. There is no a priori hypothesis about $f$ and it should be treated as a black-box entity. 

Heuristic algorithms offer guidance based on the problem-domain knowledge for optimizing function $f$. In order to design a new heuristic algorithm, having a team  of  human  experts with a  longstanding  experience  within  the  specific  domain is necessary. Usually, this is a very complex process performed with trial and error.  As a consequence, the development of hyper-heuristics which do not take advantage of the problem structure accelerated~\cite{6805577,7101236}. They are high-level methods that operate on the search space of heuristics rather than of solutions. Over the last decade, a variety of hyper-heuristic approaches have been proposed to automate the development of optimization methodologies on their own without having to rely on researchers' expertise~\cite{burke2019classification}. 
\par
The majority of popular hyper-heuristics are deployed according to the basic components of the hand-designed evolutionary algorithms~\cite{nguyen2019genetic}. We often have a population of candidate solutions that strive for survival and reproduction. However, there are no clear guidelines on the strengths and weaknesses of the proposed components that arise in other research filed such as machine learning; for developing more enhanced optimization algorithms. In this study, we claim the mentioned contribution by porting existing deep learning models from image classification to the optimization domain.
\par
NAS is currently one of the fastest-growing topics in machine learning aims to automate the neural network architecture design for various tasks such as semantic segmentation~\cite{nekrasov2019fast}, object detection~\cite{wang2019nasfcos}, and image classification~\cite{liu2018progressive}. This optimization process has focused on discovering better modeling accuracy, building architectures with lower computational complexity, or both of them. Search space, search strategy (or policy), and search speed-up methods are three main components in NAS approaches~\cite{elsken2019neural}. Already by now, outstanding results have been achieved using NAS that are superior to the expert-designed architectures~\cite{zoph2018learning}. 

Motivated by the recent successes of NAS, we propose to extend NAS studies to find an efficient neural network for the problems that arise in the global optimization domain. That is, we build and train neural networks to efficiently and adaptively solve optimization problems.  The critical contribution of this study is to tackle two major challenges that are known to the direct application of NAS for stochastic optimization. First and foremost, the widely-used search speed-up methods in NAS such as parameter sharing~\cite{liu2018darts} might not be suitable for the optimization problems. Although the parameter sharing avoids training each architecture from scratch, it would lead to unstable and sub\-optimal solutions as discussed in~\cite{li2019stacnas}. In some cases, this could end up with solutions with worse performance compared to the conventional methods in the literature; considering the complex
and  highly  non-linear optimization problems. Second, the existing NAS methods introduce a domain-specific bias with search spaces tailored to particular applications~\cite{elsken2019neural,zhou2019auto}. For example in computer vision, NAS is defined to search for the convolutional and fully connected layers  in convolutional neural networks (CNNs); without fine-tuning the hyperparameters in the learning algorithm. However, this may prevent finding superior architectures that go beyond the classification tasks. Integrating prior knowledge about typical properties of optimization problems can characterize the complexity of the search space more properly and simplify the search. Even so, the search space is huge and may contain more than $10^{15}$ different architectures~\cite{zoph2018learning}.


Particularly, we are interested to adopt the proposed NAS methods from the computer vision domain. These deep neural models use a hierarchy  of  features in conjunction with several layers to learn complex non-linear mappings between the input and output layers. As opposed to traditional methods that use handmade features, the  important features are discovered automatically  and  are  represented  hierarchically. This is known to be  the   strong   point   of  CNNs  against traditional approaches. Accordingly, these models have been described as universal learning approaches that are not task-specific and can be used to tackle different problems that arise in different research domains~\cite{alom2018history,rakhshani2019feature}. They are a regularized version of fully-connected neural networks inspired by biological visual systems~\cite{krizhevsky2012imagenet}. The "fully-connectedness" of CNNs enables them to tackle the over-fitting problem and it is reasonable to postulate that they may outperform classical neural networks for difficult optimization tasks~\cite{rakhshani2019feature}.
\par
Several works introduced deep neural networks to find the optimum solution for the optimization tasks. DNGO~\cite{snoek2015scalable} uses deep neural networks for hyperparameter optimization of large scale problems with expensive evaluations. The key point is to take the advantages of large-scale parallelism to provide an approximate model of the real objective function. The scalability of DNGO is successfully verified against the Gaussian process. Moreover, OptNet~\cite{amos2017optnet} proposed to learn optimization tasks by incorporating deep networks, bi-level optimization, and sensitivity analysis. In another study, researchers put forward MaNet optimization algorithm based on the CNN models~\cite{rakhshani2019feature}. MaNet uses feature selection to skip irrelevant or partially relevant information and uses those which contribute most
to the overall performance. The experiments indicate that MaNet is able to yield competitive results compared to one of
the best hand-designed algorithms for the CEC $2017$ problems~\cite{wu2017problem}, in terms of the solution accuracy and scalability.

Overall, all these studies make it likely that an optimization algorithm based on neural networks can find solutions that substantially outperform the state-of-the-art optimization methods. We thus found it important to go beyond hand-designed algorithms for optimization tasks by applying NAS to this less explored domain. Among different models, NAS equipped with CNN models is used in this research. CNN is a deep learning model that interleaves convolutional layers to filter redundant or even irrelevant input data to increase the performance of the network~\cite{he2016deep}. This consideration also reduces the dimensionality of the input data and  speeds up the learning process in the CNNs. Besides, it allows CNNs to be deeper networks with fewer parameters.
\par
The proposed method has several moving parts including CNN's search space, search strategy, and search acceleration. We
show how these different components could be adopted to achieve  better performance within a limited computational time. Finally, we claim that NAS can be used in other research fields and encourage further works in this domain. Our contributions can be summarized as follows:

\begin{itemize}
  \item Deep neural networks are formally defined tailored to global optimization. On top of that, a search strategy is used to perform NAS over a \textit{cell-based} search space. We show that the well-known NAS approaches can be further enhanced by considering the key properties of the optimization problems.
  \item A set of experiments are conducted to investigate the performance of the proposed method. Our contribution achieves competitive performances for CEC $2017$~\cite{wu2017problem} benchmarks and also protein structure optimization~\cite{RAKHSHANI2019100493} compared to the state-of-the-art hand-crafted algorithms.
  \item The transfer learning~\cite{litjens2017survey} and ensemble learning~\cite{dietterich2000ensemble} concepts are used to show how the optimization process can be accelerated.

\end{itemize}

The rest of the paper is organized as follows. Section~\ref{methodology} elaborates the proposed methodology with its technical details. In Section~\ref{experiments}, several experiments are conducted to show the advantages of deep learning over the conventional hand-designed algorithms in the literature. The obtained results are discussed in Section~\ref{discussion}, followed by a conclusion in section~\ref{conclusion}.

\section{Methodology\label{methodology}}
Our immediate aim is to examine the properties of the deep neural networks on optimization problems; as illustrated in Algorithm~\ref{algo0}. Such an implementation involves search space definition, search strategy, and search speed-up in parallel on GPUs. We combine the efficiency of multi-GPU systems with NAS to balance computational efficiency and the solution quality. Due to its distributed nature, we can deploy large-scale numbers of deep networks while learning different problems. Empirically, we show that the introduced method obtains better results with reductions in search complexity. The proposed methodology related to several prior works, mainly including DARTS~\cite{liu2018darts}, NAS-Bench-101~\cite{ying2019bench}, ASHA~\cite{li2018massively}, randomNAS~\cite{li2019random} and ~\cite{ying2019enumerating}. The performance of the introduced method confirms concerns raised in this study that state-of-the-art results can be obtained by using deep learning models. In  the  following, it  is  assumed  that  the  reader  is  familiar with the basic concepts of optimization algorithms, CNNs, and artificial neural networks. 

\subsection{Preliminary}
The state-of-the-art NAS methods can be parametrized  by: (i) search space; (ii) search strategy; and (iii) search speed-up methods~\cite{elsken2019neural}. For simplicity, we review some ubiquitous approaches regarding each aspect. Note that we are interested in CNN models, and so approaches about recurrent neural networks are out of the scope of this work.

\par 

\begin{figure}[ht]
    \centering
    \includegraphics[width=0.75\linewidth]{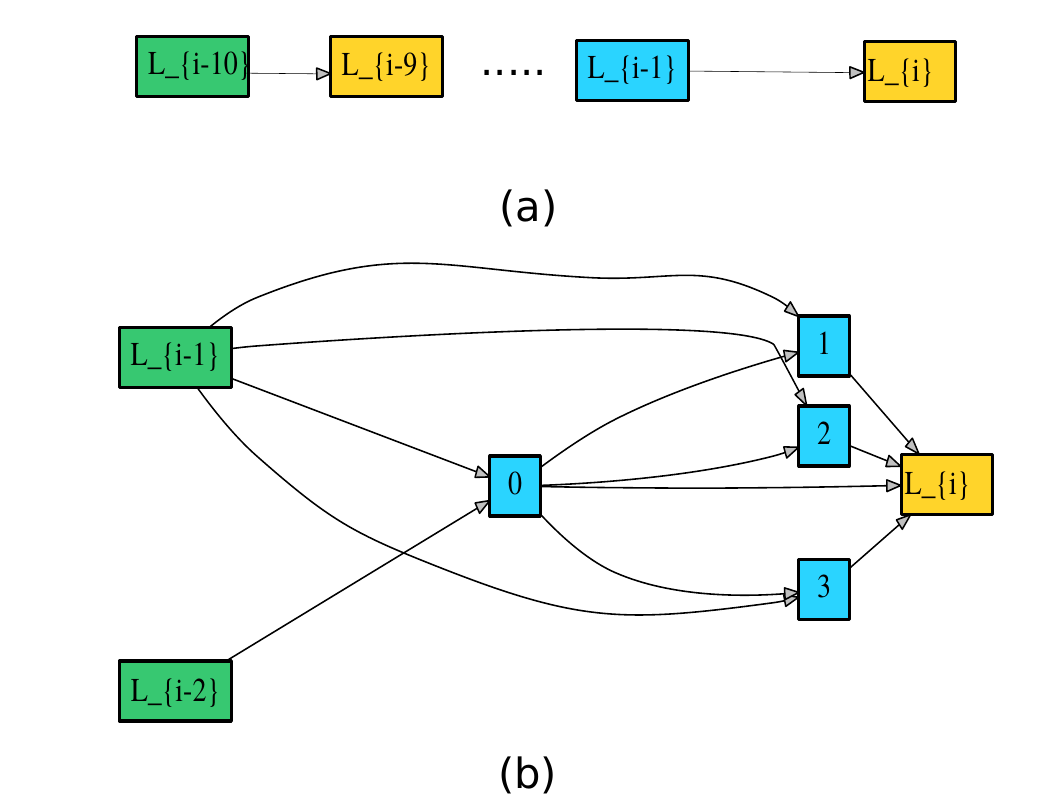}
    \caption{Illustration of the \textit{linear-structured} (a) and \textit{multi-branch} (b)  architecture search  spaces.  In the first case, NAS only allows data to flow in one direction: from a lower-numbered layer $L_{i}$  to a higher numbered layer $L_{i+1}$. The \textit{multi-branch} methods, however, allow us to use multiple branches and skip connections. Note colors are used to represent different kinds of operations at each layer.}
    \label{figure:dart}
\end{figure}

\textbf{Search space} aims to define the feasible  neural  architectures based on the applications and computation requirements. In \textit{linear-structured} NAS~\cite{cai2018efficient}, search space is defined as a sequence of $m$ layers, so as the input of the $L_i$ layer is fed by previous layer $L_{i-1}$  (Fig.~\ref{figure:dart} top). Accordingly, the number of layers, type, and hyperparameters associated with each layer will form the possible architectures' search space. \textit{Multi-branch} methods, on the other side, allow the researchers to build complex architectures with significantly more degrees of freedom. These methods are motivated by the Residual networks~\cite{he2016deep} and DenseNets models~\cite{huang2017densely} which introduced skip connections (Fig.~\ref{figure:dart} bottom). 

    



    



    



\begin{algorithm}[ht]
\SetAlgoLined

   Initialize the population with random neural networks topology
   
   Use each network to optimize $f^*$

    \Repeat{a termination condition is satisfied}{
    
      Generate new neural networks
      
      Evaluate new models to optimize f*
      
      Select models for the next iteration
    }
 \caption{The proposed method for optimizing objective $f$}
 \label{algo0}
\end{algorithm}
\vspace{0.5cm}
In the same direction, \textit{cell}-based approaches aim  to  formulate the search space by stacking several copies of  the  discovered \textit{cells}, which significantly reduced the size of the search space since \textit{cells} have fewer layers than final architectures. In \textit{micro search}, the whole architecture is built by stacking the \textit{cells} in a predefined
manner~\cite{zhang2017mixup}, while in \textit{macro search} they can be combined arbitrarily~\cite{cai2018path}.

\textbf{Search strategy} is then used to explore the above-mentioned space of neural architectures. Typical NAS approaches apply reinforcement learning~\cite{zoph2016neural}, evolutionary algorithms~\cite{elsken2018efficient}, Bayesian optimization~\cite{kandasamy2018neural}, and random search~\cite{real2019aging} to reduce the computational costs, improve the performance, or obtain a trade-off.
 
\textbf{Search speed-up strategies} lead to accelerated NAS methods for training the neural architectures, which sometimes need thousands of GPU days for NAS~\cite{zoph2016neural}. The lower fidelity~\cite{real2019aging}, learning curve~\cite{baker2017accelerating}, weight inheritance~\cite{elsken2018efficient}, and weight sharing~\cite{xie2018snas} are among the most recent approaches.

\subsection{Search space}
The \textit{input layer}, \textit{intermediate layers} (e.g., convolution layer, a pooling layer, a Dense layer, etc.), and \textit{output layer} are the basic building blocks of the CNN models. The inputs to \textit{intermediate layers} are fed by a previous layer, thus forming a network. It is necessary to define the topology of this network before deploying a CNN model in the context of the optimization problem we are trying to solve. Some network structures might lead to the final networks that are highly memory demanding and time-consuming~\cite{gu2018recent}. The key point is to define a search space that makes optimal use of computational resources and reduces the probability of generating sub-optimal network architectures.

\par

\begin{figure}[!htbp]
    \centering
    \includegraphics[width=1.0\linewidth]{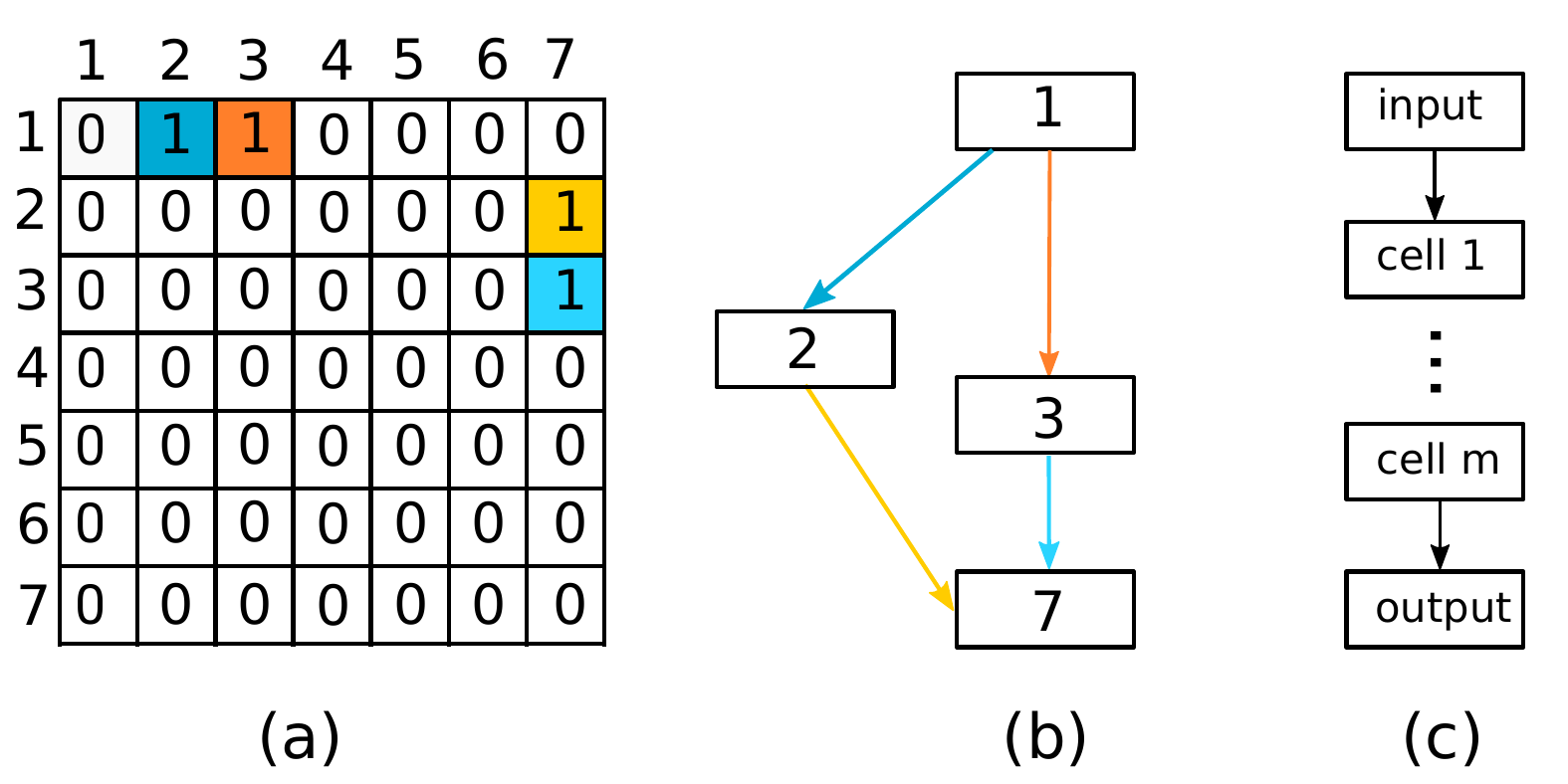}
    \caption{An overview of the \textit{cell}-based search schema: (a) The adjacency matrix used to represent the DAG. Here, $1$ and $7$ indexes belong to the \textit{input} and \textit{output} layers of the defined \textit{cell}, followed by the \textit{intermediate nodes} $2$-$6$, (b) The obtained \textit{cell} structure, where operations at each vertex are denoted by a different color, and (c) The derived final architecture with $m$ \textit{cells}.}
    \label{figure:dag}
\end{figure}

Following~\cite{liu2018darts}, we limit the NAS search space  by factorizing each architecture into  \textit{multi-branch} \textit{cells}. This representation is characterized by the number of layers $m$, choice of operation $O=[o_1, o_2, ..., o_n]$ for each layer $L$, and ${\theta}_{L_{o}}$ which denotes the associated hyperparameters for the operation $o$ at the $L$-th layer. At a higher level, the entire network is defined by $m$ \textit{cells} connected sequentially, which can be seen in Fig.~\ref{figure:dag}. A \textit{cell} is composed of one \textit{input layer}, several intermediate nodes, and one \textit{output layer}. Specifically, the \textit{input layer} is connected to the output of the previous \textit{cell} layer. Moreover, the \textit{output layer} aggregates the representations from all the intermediate nodes. The  layouts  of the intermediate nodes can be defined using a directed acyclic graph (DAG), where a node contains the results from a previous operation and an edge $e_{ij}$ shows some operation $o$ that transforms the feature map from node $I_i$ to $I_j$.  Thus we have intermediate nodes:
\begin{equation}
    I_{j} = \sum_{i<j} o_{i,j}(I_i)
\end{equation}

where $o_{i,j}$ denotes the selected operation from lower indexed node $i$ to higher node $j$. At each layer $L$, one of three possible operations convolution, max pooling and average pooling from CNN models can be chosen. In the following, each of these layers is characterized to fully parameterize the neural architecture space.

A typical CNN network is composed of a series of convolutional and pooling layers. Generally speaking, convolutional layers provide a way of capturing the dependencies in their input by applying different filters.

For a given filter and the input data, the convolution operation takes entries with size $p \times p$ of the input and multiplies by the filter. The sum of the entries is then the first entry of the so-called feature map. The weights of the filters are adopted during the training process, while the number of filters and their size should be configured. Given the filter weights, we create a sliding window $p \times p$ that goes by step size $s$ through the vertical and horizontal dimensions of the input data. The hyperparameters of convolutional layers are: a) the number of filters $n \geq 1$, b) size of the sliding window $p \times p \geq 1$, and c) the stride step size $s \geq 1$. For each filter, a fixed weight will be used across the entire input. A one layered CNN with $n=10$ filters of size $5 \times 5$ and $10$ biases has $5 \times 5 \times 10 + 10 = 260$ parameters, while a fully connected network for a $K = (P \times M)$ image with $250$ neurons has $(250 \times K+1)$ parameters. This is the main advantage of CNNs which makes them more efficient in terms of memory and complexity; compared to fully connected neural networks.
\par

\vspace{0.5cm}
Similar to convolutional layers, pooling provides another way to reduce the dimension of a layer. Although they can be replaced by the convolutional layers, they provide a simpler way by summarizing a $p \times p$ area of the input with certain fixed weights. For an average pooling layer with $n$ feature maps, we should have a convolutional layer with $n$ filters of size $p \times p$ and stride $s$. The $i$th filter has the values as in Eq.~\ref{eq_2} for the dimension $i$, and zero for the other dimensions $i \in [1,n]$.

\begin{equation}
\begin{pmatrix}
\frac{1}{p^2} & \cdots  & \frac{1}{p^2}\\ 
\vdots  & \ddots  & \vdots \\ 
\frac{1}{p^2} & \cdots & \frac{1}{p^2}
\end{pmatrix}
\label{eq_2}
\end{equation}

\begin{figure}[!htbp]
    \centering
    \includegraphics[width=1.0\linewidth]{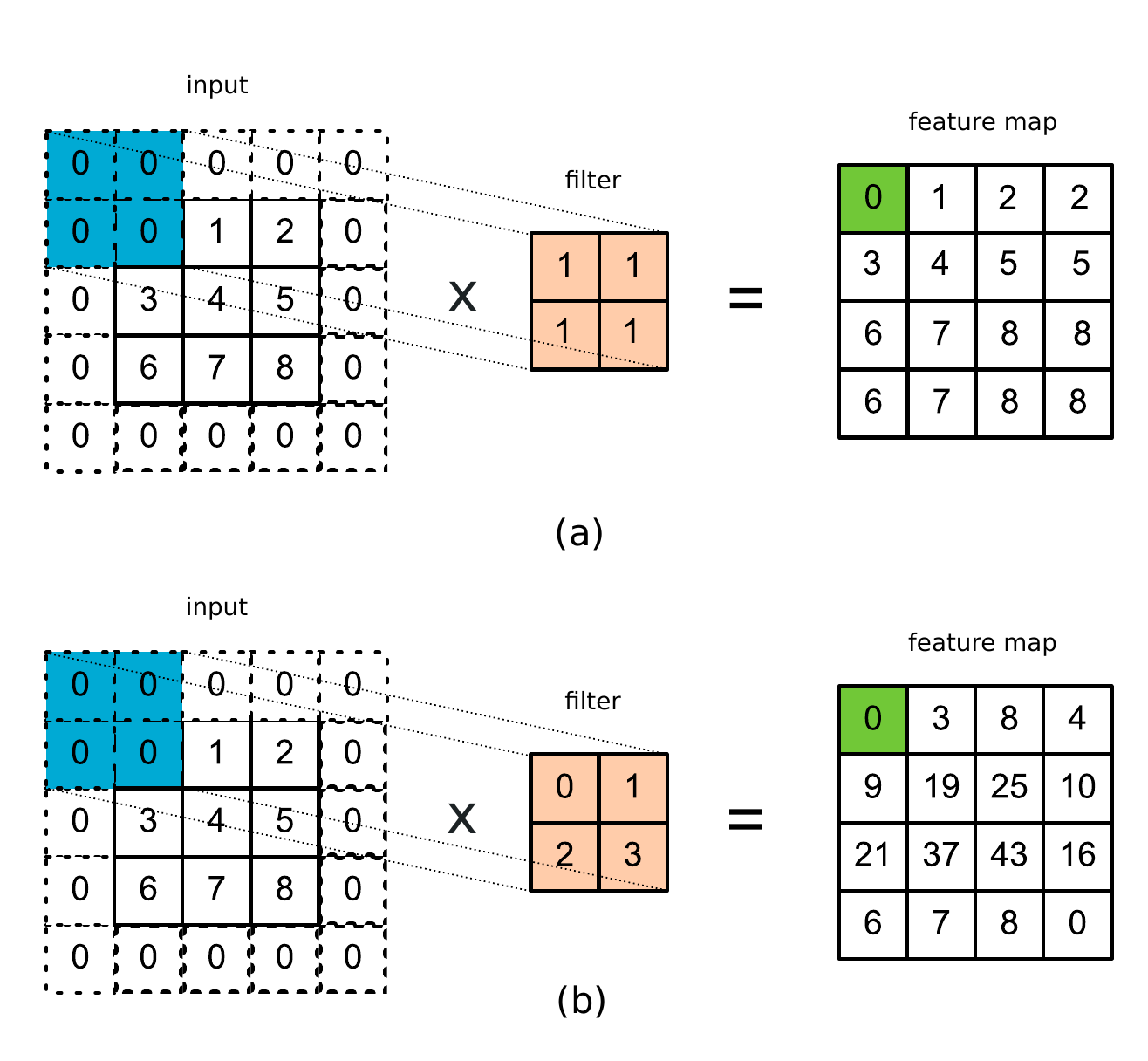}
    \caption{Application of a single max pooling layer (a) and a convolutional layer (b) with $1$ filter of size $2 \times 2$ with stride $s = 1$ to input data of size $5 \times 5$. In (a), the maximum value after element-wise multiplication is taken, while in (b) the summation is computed. This figure also shows how the max pooling layer can be replaced by a convolutional one.} 
    \label{fig:conv_layers}
\end{figure}

The same thing can be considered in max pooling, where the maximum value within the window is taken with filter weights $1$. It is not necessary to pool over the whole input and we can pool over a window with stride step size $s$. So, we have only two hyperparameters $p$ and $s$ for the pooling layer. To sum up, the functionality of a simple convolution layer and max pooling layer is depicted in Fig.~\ref{fig:conv_layers}.

Besides the mentioned operations, the CNN learning algorithm itself contains a set of hyperparameters. The batch size is a hyperparameter of the learning algorithm that controls the number of data that will be propagated through the network. The appropriate batch size can increase the accuracy of the learning algorithm when training a neural architecture. We found that fine-tuning this hyperparameter can have a significant impact on the performance of the NAS (we will show that this consideration is preferred over traditional NAS methods that use a fixed value). So, two typical choices of batch size $\in \{1, 32\}$ are used in this study.

We now quantify the size of our search space to determine the magnitude of the proposed NAS method. A comparison of the search space complexity for state-of-the-art NAS methods is given in Table~\ref{table:complexity}. The space of the $cell$ networks contains all DAG graphs on $v$ nodes, where each node denotes one operation with $p=3$ and $s=1$. In this work, the number of operations is limited to: a) one convolutional layer, b) one max pooling layer, and c) one average pooling layer. Moreover, the maximum number of nodes in each \textit{cell} is supposed to~be~$\leq 7$. Also, the maximum number of edges is limited to $9$. Considering $21$ possible edges in DAG adjacency matrix, $3$ operations for each intermediate node, and $2$ different values for batch size, $2^{21}~\times~3^{5}~\times2\approx~1.0 \times 10^9$ total models exist in this search space. The created models do not apply \textit{ReLU} activation function or batch normalization between depthwise and pointwise layers. To match shapes in convolutional layers, strided $1 \times 1$ convolution projections are applied as necessary. Furthermore, the output of the intermediate blocks is concatenated. 

\subsection{Problem formulation}
We aim to find a topology that minimizes a considered objective function $f(x)$ over a neural search space $\mathcal{D}$ with the available computational budget $\mathcal{T}$. Formally, this is
equivalent to search for a superior neural architecture $\mathcal{A}^{*} \in \mathcal{D}:$

\begin{equation}
    \mathcal{A}^{*} = arg\min_{\mathfrak{A} \in \mathcal{D}} cost(\mathfrak{A}, f, w, \mathcal{T}) + \xi
\end{equation}

\begin{table}[!hb]
\centering
\caption{Comparisons of the search space complexity between the introduced  and state-of-the-art NAS methods.}
\begin{tabular}{l|l|l}
\midrule
\makecell{Search Method}       & \makecell{Number of Layers} & \makecell{Search complexity}         \\ \midrule
EDNAS~\cite{lee2019efficient}  & $6$                & $1.04 \times 10^9$ \\
PNAS~\cite{liu2018progressive}   & $5$                & $10^{12}$              \\
NASNet~\cite{zoph2018learning} & $5$                & $10^{28}$              \\
RENAS~\cite{chen2019renas}       & $5$                & $3.1 \times 10^{13}$ \\
EPNAS~\cite{perez2018efficient}               & $5$                & $5 \times 10^{14}$   \\
STACNAS~\cite{li2019stacnas}                & $4$                & $10^{18}$              \\ \midrule
current study                & $5$                & $1.0 \times 10^9$              \\ \midrule
\end{tabular}
\label{table:complexity}
\end{table}

where \(w\) is the learned weights of \(\mathfrak{A}\) and \(\xi\) is a penalty function. The measure of violation in $\xi$ is nonzero when the number of edges $\vartheta$ in the DAG graph is $>9$ or when there is no path from the input to the output layer; and is zero in the other cases. The mathematical function associated with $\xi$ is:
\begin{equation}
 \xi = \left\{\begin{matrix}
(\vartheta-9) \times \eta_{1}  & \textit{if } \vartheta>9 & \\ 
\kappa  \times \eta_{2} \hspace{24pt} & \textit{\hspace{12pt} otherwise }& \\
\end{matrix}\right.
\label{eq4}
\end{equation}

In Eq.~\ref{eq4}, $\kappa$ denotes the number of single nodes and $\eta_{2},\eta_{1}$ are the penalty coefficients. Before applying any
search regime, the representative DAG graph for arbitrary $\mathcal{A}$ network should be encoded in its genotype form. We adopt a very general encoding: the first $21$ binary genes $\in \{0,1\}$ are used to represent the edges in the graph, while other $5$ genes $\in\{0,1,2\}$ are used to represent the type of the operation. Also, the last gene denotes the batch size hyperparameter.

\subsection{Search strategy}

Up to our best knowledge, this paper describes the first attempt to extend NAS for the optimization domain. Accordingly, 1) random search, 2) reinforcement learning, and 3) Bayesian optimization strategies are used to leverage the observations of previous studies, respectively. As shown in Algorithm~\ref{algo1}, they all operate in a similar fashion: in each iteration $i$ a step vector $\Delta \mathfrak{A}$ is computed by means of an updated formula $\varpi$. Thereafter, this formula is updated using $\Phi$ to guide the search process more effectively. For the reinforcement learning and Bayesian methods, $\Phi$ utilizes some  history of the generated architectures and their associated performance evaluated at the current and past iterations. For example, the update formula is based on recurrent neural networks in the second method, while reinforcement learning is used to update the aforementioned network according to the past information.

\textbf{1-Random search} is the most simple yet effective~\cite{li2019random} baseline in this study. We used our implementation according to which the generated candidates are drawn from a uniform probability distribution and are independent of the samples that come before it. These properties make it well suited to highly parallel systems. Moreover, random methods are flexible in that they can be applied to both the continuous and discrete search space; in contrast to Bayesian approaches based on Gaussian processes~\cite{klein2016fast} and gradient-based approaches~\cite{liu2018darts}. 
 
\textbf{2-Reinforcement learning} is another approach that is proposed to search for good architectures. Following~\cite{zoph2016neural}, a recurrent network trained by reinforcement learning is used to generate better architectures; as time goes on.

\begin{figure}[!ht]
    \centering
    \includegraphics[width=0.85\linewidth]{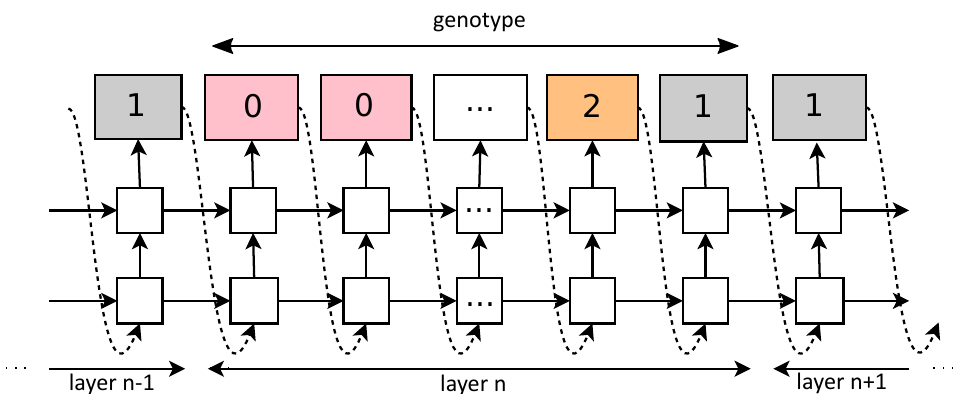}
    \caption{Application of a recurrent neural network for sampling a new $cell$ topology~\cite{zoph2016neural}. The generated networks in the next time steps are influenced by what the network has learned from the past.}
    \label{figure:rnn}
\end{figure}

\begin{algorithm}
\setstretch{1.2}
\SetAlgoLined
\KwIn{NumIterations, SearchSpace, objective $f$}
\KwOut{BestCost}
   $\mathfrak{A}_1 \gets \textnormal{RandomSolution(SearchSpace)}$
   
   \For{$i \in \{2,..\textnormal{NumIterations}\}$}{
   
    $\textnormal{Cost}_{i-1} \gets
    \textnormal{BestSolution}(\mathfrak{A}_{i-1}, f)$
   
    \begin{tikzpicture}
        \node[fill={rgb:black,1;white,2}] (A) [draw, rectangle] {%
            \begin{varwidth}{\linewidth}
                $\Delta \mathfrak{A} \gets \varpi(\left \{ \mathfrak{A}_j, f \right \}_{j=1}^{i-1}) $
        \end{varwidth}
            };
        \node[fill={rgb:orange,1;yellow,2;pink,5}] (B)[draw, rectangle, anchor=west,xshift=8pt] at (A.east) {%
            \begin{varwidth}{\linewidth}
                $\varpi(.)=\left\{\begin{matrix*}[l]
                \textnormal{1-random distribution}\\ 
                \textnormal{2-recurrent network}\\ 
                \textnormal{3-feature selection}
                \end{matrix*}\right.$
            \end{varwidth}
            };
    \path[->,>=Latex] (A) edge (B);
    \end{tikzpicture}

    $\mathfrak{A}_i \gets \mathfrak{A}_{i-1} + \Delta \mathfrak{A}$
    
    \begin{tikzpicture}
        \node[fill={rgb:black,1;white,2}] (A) [draw, rectangle] {%
            \begin{varwidth}{\linewidth}
                $\varpi \gets \Phi (\varpi,\mathfrak{A}_i, f) $
        \end{varwidth}
            };
        \node[fill=blue!20] (B)[draw, rectangle, anchor=west,xshift=7pt] at (A.east) {%
            \begin{varwidth}{\linewidth}
                $\Phi(.)=\left\{\begin{matrix*}[l]
                \textnormal{1-not applicable}\\ 
                \textnormal{2-reinforcement learning}\\ 
                \textnormal{3-surrogate model}
                \end{matrix*}\right.$
            \end{varwidth}
            };
    \path[->,>=Latex] (A) edge (B);
    \end{tikzpicture}
    }
	$\textnormal{BestCost} \gets \min(\textnormal{Cost})$
 \caption{The General structure of the NAS methods}
 \label{algo1}
\end{algorithm}
 The recurrent networks are able to use information learned from prior steps while generating new architectures. Let's suppose we would like to search for a $cell$ topology, the agent’s action is to generate new architecture, while its reward is based on the performance of the trained architecture on our optimization problem. In Fig.~\ref{figure:rnn}, it is shown how a recurrent network can be used to sample new architectures as a sequence of tokens. The list of the predicted actions $a_{1:T}$ by the controller will be used to generate a new architecture. This is the key feature of recurrent nets which allows us to operate over sequences of vectors in the input, the output, or both.\footnote{For more details please see: \url{http://karpathy.github.io/2015/05/21/rnn-effectiveness}}. Hence, the controller should only maximize its expected reward $J(\theta_c)$ which directly depends on the parameters of the recurrent net $\theta$; as presented in~\cite{williams1992simple}. We can use reinforcement learning as follows:

\begin{equation}
    \bigtriangledown_{\theta_c} J(\theta_c)=\frac{1}{M}\sum_{k=1}^{M}\sum_{t=1}^{T} \bigtriangledown_{\theta_c} log P(a_t|a_{t-1:1};\theta_c)R_k
    \label{eq5}
\end{equation}

In Eq.~\ref{eq5}, $M$ denotes the number of sampled architectures in one batch, $T$ is the dimension of the problem, and $R_k$ is the performance of the k$^{th}$ neural network architecture after being trained.

\textbf{3-Bayesian methods} are widely used for hyperparameter optimization, but their application to NAS has been limited since they mainly employ the Gaussian process and they are not appropriate for high dimensional NAS~\cite{elsken2019neural}. In this work, we adopt a single-objective version of MAC \cite{10.1007/978-3-030-12598-1_20} as the Bayesian search strategy for NAS. Accordingly, we generate a set of random architectures in the early iterations in parallel, while we adopt a feature selection strategy to generate more promising candidates in the later stages of development. This is a key property in MAC that helps us to make a balance between the solution quality and the computational time.~\footnote{The application  of MAC  to  very recent  benchmarks  against  Random search, SMAC, and F-RACE state-of-the-art  competitors,  at  most contributes to top performances. These results encourage us to utilize MAC in this study.}
 
After half of the iterations, a response surface model is created to provide a fast approximation of the expensive evaluations for the later stages of evolution. Given a set of configurations \(\mathfrak{A}_{1},...,\mathfrak{A}_{N} \in \mathbb{R}^{\mathcal{D}}\) with known performance $\mathbf{y}$, the radial basis function (RBF) interpolant for $\hat{\mathfrak{A}}$ is then computed as below~\cite{doi:10.1287/ijoc.1060.0182}:
\begin{equation}
\tilde{y}({\hat{\mathfrak{A}}})=\sum_{i=1}^{N} \lambda_{i}\phi( \begin{Vmatrix} \hat{\mathfrak{A}} - \mathfrak{A}_{i} \end{Vmatrix} ) + p(\hat{\mathfrak{A}}), \ \hat{\mathfrak{A}} \in \mathcal{D}
\end{equation}

Here, $\mathfrak{A}_{i}$ is the associated solution representation for \textit{i-}th architecture model. Also, $\begin{Vmatrix}.\end{Vmatrix}$ is the Euclidean norm, $\lambda_{i} \in \mathbb{R}$ for $i=1,...,N$, $p \in  \prod_{m}^{d}$ denotes the linear space of polynomials in $d$ variables  of  degree which is less  than  or  equal  to $m$, and $\phi$ is a RBF kernel. Following \cite{doi:10.1287/ijoc.1060.0182}, MAC selected the \textit{surface splines} $\phi(r)= r^k$ form with $k=3$ as the RBF. Having this in mind, we can compute a matrix $\Im \in \mathbb{R}^{N\times N}$ by $\mathbf{\Im}_{i,j}= \phi( \begin{Vmatrix} \mathfrak{A}_{i}- \mathfrak{A}_{j} \end{Vmatrix} )$; $ i,j= 1,...,N$. Assume that $\hat{m}$ is the dimension of the linear space  $\prod_{m}^{d}$ such that $m \geq =\lfloor k/2 \rfloor$. Accordingly, we have another matrix $\mathbf{P} \in \mathbb{R}^{n \times \hat{m}}$ such that: $P_{ij}=p_{(i)}(\mathfrak{A}_{(i)})$, $i=1,...,N; j=1,...,\hat{m}$. The approximated model can then be obtained by solving the system as presented in Eq.~\eqref{eq011}, where $\mathbf{c}={(c_{1},...,c_{\hat{m}})}^{T} \in \mathbb{R}^{\hat{m}}$.

\begin{equation}
\mathbf{}
\begin{pmatrix}
\mathbf{\Im}  & \mathbf{P}\\ 
\mathbf{P}^{\top}& 0
\end{pmatrix} \begin{pmatrix}
\mathbf{\gamma}
\\ 
\mathbf{c}
\end{pmatrix}=\begin{pmatrix}
\mathbf{y}
\\ 
0_{\hat{m}}
\end{pmatrix}
\label{eq011}
\end{equation}

After half of the iterations, the randomly generated perturbations $\rho$ are  also ranked  in descending  order  according  to  their  contribution to the validation accuracy:  the  top  50\%  with  \textit{promising} label and  the  others with \textit{non-promising} label. Given this training set, MAC applies a feature selection strategy to  obtain weight for generating the new configurations; as elaborated in~\cite{10.1007/978-3-030-12598-1_20}.

\subsection{Search speedup}

 The above mentioned architecture $\mathfrak{A}_{i}$ should be trained on each function $f$ which can take a very long time. However, we would like to solve our optimization task with roughly the same wall-clock time needed for an evolutionary algorithm. To tackle this challenge, we help ground the used methods by introducing two considerations: 1) early-stopping  and 2) identifying isomorphic computational graphs. In the first case, we stop training the model once its performance stops improving on the fitness function $f$. Accordingly, we define a threshold $\psi$ to consider whether a function value at some epoch as improvement or not. If the difference of improvement compared to the previous training epoch is below $\psi$, it is quantified as no improvement. This very simple paradigm can prevent the architecture to spend the computational time on sampling in non-optimal regions. Moreover, it only  needs  a  few  modifications and is easy to implement. In particular, the idea is to 1) assign a small computational budget $r$ to the sampled architecture, 2) train the architecture, 3) increase the budget for architecture by a factor of $\eta$, and repeat until the maximum budget of $\iota$ for the architecture is reached or its performance stops improving.
 
 The second consideration is inspired by~\cite{ying2019bench, ying2019enumerating} according to which we check isomorphic computational \textit{cells} before evaluating the generated model. Of course, there is no guarantee to perfectly find all pairs of non-isomorphic $cells$, but this will work in many cases. The key point is to reduce the size of the search space by detecting the \textit{cells} which have different genotypes but encode the same computation. This strategy could significantly reduce the size of our defined search space. Moreover, it can  help the RL and MAC from being misled to a false optimum by using the information from the model.

\subsection{Putting it all together}
Already by now, we described how NAS can be applied to generate different architectures within the defined search space. In this subsection,  we would like to  put  all  of  the details together to show how such encoded architecture $\mathfrak{A}$ can be employed to solve an optimization problem. Since we assume that not all readers are fully familiar with machine learning, we used the optimization terminologies as a common language to describe all the aspects. 

The first principle is that in CNNs you present your input image and train your model to make predictions on unseen data. This implies that our population is represented by a set of $NumSol$ random $n \times n$ inputs for the model (i.e., the raw pixel values of the image).  So, as opposed to evolutionary algorithms, each solution $x \in \mathbb{R}^{D}$ is represented by a matrix with any arbitrary size $n \times n \geq D$ rather than a vector. During the training of the network, the defined convolutional operations transform this $n \times n$ matrix, layer by layer, to a final feasible solution $x \in \mathbb{R}^{D}$. This large-part genotype representation enables the optimizer to keep genetic information that was necessary for the past as a source of exploration, as well as a playground for extracting new features that can be advantageous in the exploitation. The second important point to note is that CNNs only modify their weights and the input data are kept fixed, while evolutionary algorithms modify their initial population. They adopt gradient descent to update these weights based on the backpropagation of the error algorithm. Here, the distance from optimal $f(x^*)~-~f(x)$ is used to calculate the model error. 

The training process of a neural network using Adam optimizer~\cite{AdamRef} for an objective function $f(x)$ is summarized in Algorithm~\ref{algo2}. These steps are simplified and we just tried to provide intuition into the training process. In Algorithm~\ref{algo2}, $\Delta_w E_{t}{(w_{t-1})}$ denotes the partial derivatives of $E_{t}$ with respect to $w$ at time step $t$, $\alpha$ is the learning rate, and $\beta_1$, $\beta_2$ hyper-parameters are the exponential decay rates of the $m_t$ and $v_t$ moving averages, respectively~\cite{AdamRef}. 

\begin{algorithm}[!htbp]
\setstretch{1.1}
\SetAlgoLined
\KwIn{NumSol, n, $f$, $\alpha$, $\beta_1$, $\beta_2$, $\mathfrak{A}$}
\KwOut{Best}

   $m_0 \gets 0$, $v_0 \gets 0$, $t   \gets 0$
   
   $w_t \gets \textnormal{RandomWeights} (\mathfrak{A})$
   
   $\textnormal{Best} \gets inf$
   
    \Repeat{stopping criteria are met}{
    
      $x \gets \mathfrak{A}(w_t, NumSol, n)$
      
      $\textnormal{Best} \gets \min(\textnormal{Best}, f(x_i)); i=1,...,\textnormal{NumSol}$
      
      $t \gets t+1$
      
      $E_{t}{(w_{t-1})}=\frac{1}{\textnormal{NumSol}}\sum_{i=1}^{\textnormal{NumSol}} f(x_i)-f(x^*)$
      
      $g_t \gets \Delta_w E_{t}{(w_{t-1})}$
      
      $m_t \gets \beta_1 \times m_{t-1} + (1 - \beta_1) \times g_t$
      
      $v_t \gets \beta_2 \times v_{t-1} + (1 - \beta_2) \times g_{t}^{2}$
      
      $\hat{m}_t \gets \frac{m_t}{(1 - \beta_1^{t})}$
      
      $\hat{v}_t \gets \frac{v_t}{(1 - \beta_2^{t})}$
      
      $w_{t} \gets w_{t-1} - \frac{\alpha \times \hat{m}_t}
      {(\sqrt{\hat{v}_t} + \epsilon)}$
    }
 \caption{Optimizing objective $f$ using Adam}
 \label{algo2}
\end{algorithm}

 Let's see how forward propagation step can be used to generate a $9$-dimensional solution vector $x$. Suppose that we have a model with input matrix $n \times n = 5 \times 5$, $NumSol=1$, one convolution layer, and one max pooling layer. If we use a $k \times k = 2 \times 2$ filter with stride size $s=1$ and padding size $p=0$, the output of the first layer will be of size $o(\mathfrak{A},conv) = \frac{n-k+2p}{s} + 1 = \frac{5-2+2 \times 0}{1} + 1 = 4$. Thereafter, we apply the max pooling with the same hyperparameters and we have  $o(\mathfrak{A},max) = \frac{o(\mathfrak{A},conv)-k}{s} + 1 = \frac{4-2}{1} + 1 = 3$. We can see how the neural network $\mathfrak{A}$ is used to reduce the dimensionality of the input; $\mathfrak{A}: \mathbb{R}^{5 \times 5} \rightarrow  \mathbb{R}^{3 \times 3} $. The output of the max pooling layer which is $x \in \mathbb{R}^{9}$ vector then forms our genotype for computing the error value.

\section{Experimental Results\label{experiments}}

In   this   section, we conduct a pipeline  of  experiments to answer the following questions.

\begin{itemize}
  \item[--] How effective is the optimization with neural networks?
  \item[--] What are the influences of the different NAS search strategies?
  \item[--] Does the proposed methodology is scalable?
  \item[--] How much efficiency is gained from using a trained network to solve another similar problem?
  \item[--] How much efficiency is gained from using an ensemble network to solve another similar problem?
  \item[--] What are the influences of transferring the learned knowledge for solving several problems; to a new similar task?
\end{itemize}

 To provide a fair comparison, the same settings, computational resources and budgets are adopted for all the results. The experiments are performed by using the parallel power of graphics cards~\footnote{Operating system: GNU Linux, CPU: Intel(R) Xeon(R) CPU E$5$-$2670$ $0$ @ $2.60$GHz, Tesla K$40$c , Main memory: $16$ GB, GPU memory, $12$ GB, Programming language: Python}. 

\subsection{Experimental setup}

\textbf{Case studies:} The elaborated NAS methodology is applied to learn architecture for two different optimization problems. We conduct experiments based on CEC $2017$ benchmarks to assess the performance of the RS, RL, and MAC strategies. In the same section, the results are compared to the progressive neural architecture search (PNAS)~\cite{liu2018progressive}. We used a set of 9 particularly challenging unimodal (F$1$ and F$3$) and multimodal (F$4-$F$10$) functions~\footnote{Function F$2$ has been excluded by the organizers because it shows unstable behavior especially for higher dimensions~\cite{wu2017problem}.}. These problems are first implemented on GPU and are then linked with the TensorFlow machine learning library~\footnote{\url{https://www.tensorflow.org}}. All CEC test functions should be minimized within the search ranges $[-100, 100]^\textnormal{D}$. 

Next, the introduced method is compared with state-of-the-art hand-designed algorithms for real-world PSP sequences from protein data bank~\footnote{\url{https://www.rcsb.org}}; given in Table~\ref{table:proteins}. The  AB  off-lattice  model  is used to define the problem in continuous search space according to which we can predict the  secondary  structure  of  a  protein  using  its  amino  acid sequence. This secondary conformation is characterized by the bond angles $[\theta_2, \theta_3, \theta_4, \cdots ,\theta_{n-1}]$, where $n$ is the number of bonds and $\theta_i \in (-180, 180]$. The optimization task then is to minimize the free energy  of  a  protein sequence as:

\begin{equation}
    \sum_{i=1}^{n-2} \frac{1-cos \theta_i}{4} + \sum_{i=1}^{n-2}\sum_{j=i+2}^{n} \left [ r_{ij}^{-12} - C (\zeta_i, \zeta_j) \times r_{ij}^{-6} \right ]
\end{equation}

where $r_{ij}$  denotes the  distance  between $i$-th and $j$-th  monomer; as given in~\cite{stillinger1995collective}.

\begin{table}[htb]
\centering
\caption{The details of protein sequences used in experiments}
\begin{tabular}{l|l|l|l}
\midrule
No. & Length & PDB ID & Sequence                   \\ \midrule
$1$   & $13$        & $1$BXP   & $\text{ABBBBBBABBBAB}$              \\
$2$   & $13$        & $1$CB$3$   & BABBBAABBAAAB              \\
$3$   & $16$        & $1$BXL   & ABAABBAAAAABBABB           \\
$4$   & $17$        & $1$EDP   & ABABBAABBBAABBABA          \\
$5$   & $18$        & $2$ZNF   & ABABBAABBABAABBABA         \\
$6$   & $21$        & $1$EDN   & ABABBAABBBAABBABABAAB      \\
$7$   & $21$        & $1$DSQ   & BAAAABBAABBABABBBABBB      \\
$8$   & $24$        & $1$SP$7$   & AAAAAAAABAAABAABBAAAABBB   \\
$9$   & $25$        & $2$H$3$S   & AABBAABBBBBABBBABAABBBBBB  \\
$10$  & $25$        & $1$FYG   & ABAAABAABBAABBAABABABBABA  \\
$11$  & $25$        & $1$T$2$Y   & ABAAABAABBABAABAABABBAABB  \\
$12$  & $26$        & $2$KPA   & ABABABBBAAAABBBBABABBBBBBA \\ 
$13$  & $29$        & $1$ARE   & BBBAABAABBABABBBAABBBBBBBBBBB \\ 
$14$  & $29$        & $1$K$48$   & BAAAAAABBAAAABABBAAABABBAAABB \\ 
$15$  & $29$        & $1$N$1$U   & AABBAAAABABBAAABABBAAABBBAAAA \\ 
$16$  & $29$        & $1$PT$4$   & AABBABAABABBAAABABBAAABBBAAAA \\ 
\midrule
\end{tabular}
\label{table:proteins}
\end{table}

\textbf{Settings:} All the experiments are performed using $\mathcal{T}~=~100,000 \times D$ fitness evaluations. As it is explained before, the generated neural architectures should be trained so as to solve the problem at hand. In NAS, gradient-based Adam optimizer is used to update the networks' weights with a learning rate of $0.001$, $\eta=0.01$. The other hyperparameters are set according to~\cite{AdamRef}. For PNAS, we followed the training procedure used in ~\cite{liu2018progressive}. However, all the normalization layers are removed and the activation functions are set to None; which is the same consideration that we applied to our methodology. The parameter details of PNAS are presented in Table~\ref{table:PNAS}, while the operation space is given as follows~\cite{liu2018progressive}: a) $3 \times 3$, $5 \times 5$, and  $7 \times 7$ depthwise-separable convolutions, b) $1 \times 7$ followed by $7 \times 1$ convolution, c) identity, d) $3 \times 3$ average pooling, e) $3 \times 3$ max pooling, and finally $3 \times 3$ dilated convolution.
\par
Regarding the search methods, the maximum number of epochs to train the model is set to $200$ epochs. We used the original settings for both the RL and MAC methods. In RL, a two-layer RNN controller with $35$ hidden units is presented. The Adam with a learning rate of $0.1$, weight decay of $1.0e-04$, and momentum of $0.9$ is adopted~\cite{zoph2016neural}. The reward used for updating the controller is the mean error value that is propagated in a single batch. In MAC, the number of generated trial samples at the pre-evaluation step is $10,000$. Accordingly, MAC builds and trains a surrogate to find the most promising architecture using the surrogate modeling to replace in part the original computationally expensive solver; which is training the neural network for each candidate.

\begin{table}[!ht]
\centering
\caption{The parameter configuration of PNAS}
\begin{tabular}{l|l}
\midrule
Description & Configuration  \\ 

\midrule
Maximum number of epochs to train the model   & $200$       \\
Dimension of the embeddings for each state   & $20$       \\
Number of epochs to train the controller & $30$ \\
Number of children networks to train   & $8$       \\
Learning rate for the child models   & $0.001$       \\
Number of cells in RNN controller   & $100$       \\
Batch size of the child models   & $32$       \\
Number of blocks in each cell   & $3$       \\
Activation function  & None       \\
\midrule
\end{tabular}
\label{table:PNAS}
\end{table}

\subsection{Results on CEC $2017$}
This section reports the empirical evaluation of the NAS search methodology on $9$ standard benchmark functions from the global optimization literature. In this section, we aim to assess the  approach's  exploration performance and so the comparisons are based on the best-obtained results over 15 runs. The upper and lower bounds for all the test functions are the same and the input data to the neural networks are $500$ randomly sampled $n \times n = 32 \times 32$ matrices. We refer the reader to the original material for a description of the test functions and their properties~\cite{wu2017problem}.

\textbf{In Table~\ref{table:results-30D}}, the results for $30$ dimensions are shown. We can see that the MAC search strategy performs well for most of the problems, suggesting that a substantial speedup can be provided by the value of the additional information available from its surrogate model. Meanwhile, we can see that RS performs surprisingly well on F$8$, whereas RL outperformed the other algorithms on F$5$. Compared to PNAS, all the extensions of the introduced methodology show competitive results with nearly an order of magnitude reduction in the objective function value. We argue that the incorporated batch size hyperparameter is the main reason for this performance improvement. We further analyze this performance by conducting a Wilcoxon signed-rank test between the MAC and the other search methods. In Table~\ref{table:results-30D}, symbol '$+$' denotes that the \textit{null} hypothesis is rejected and MAC obtained a superior performance,  symbol '$-$' shows an inferior performance, and '$=$' suggests no statistical difference between the pair-wise algorithms. 

\begin{table}[htb]
      \centering
      \caption{The performance of different search methods for NAS on CEC~2017 $30$-dimensional test cases. A pair-wise comparison between the MAC search strategy and the other competitive methods is also presented using Wilcoxon signed-rank test with $\alpha=0.05$.
      }
        \begin{tabular}{l|l|l|l|l}
        \midrule
        \makecell {No\quad \quad \quad \quad \quad} & {MAC} & {RS} & {RL} & {PNAS}  \\
         \midrule
        F1       &  $1.13118e-02$     &   $1.40977e-02$    &   ${1.94292e-03}$  &   ${1.47442e+05}$ \\

         F3       &  ${1.27183e-03}$     &   $1.78565e+04$    &   $1.02218e-01$  &   ${5.95693e+04}$ \\

         F4       &  ${2.62499e-06}$     &   $1.06736e-05$    &   $7.37121e-06$  &   ${3.03500e+00}$ \\

         F5       &  $3.98197e+00$     &   $3.97984e+00$    &   ${9.94979e-01}$  &   ${1.68726e+02}$ \\

         F6       &  $2.05215e-01$     &   ${1.26378e-01}$    &   $1.07848e+00$  &   ${5.29290e+01}$ \\
        
         F7       &  $3.40319e+01$     &   $3.42406e+01$    &   $3.41101e+01$  &   ${2.13789e+02}$ \\

         F8       &  $4.97486e+00$     &   ${9.98603e-01}$    &   $6.97262e+00$  &   ${1.46752e+02}$ \\

         F9       &  ${1.70125e-11}$     &   $2.80843e-11$    &   $4.41499e+00$  &   ${6.62968e+02}$ \\

         F10       &  ${1.39714e+03}$     &   $2.05124e+03$    &   $1.97163e+03$  &   ${3.27547e+03}$ \\
         \midrule
         $+ \setminus = \setminus -$   &      &   $5 \setminus 2 \setminus 2$    &  $5 \setminus 2 \setminus 2$ &  $9 \setminus 0 \setminus 0$ \\
        \midrule

        \end{tabular}
      \label{table:results-30D}%
\end{table}%

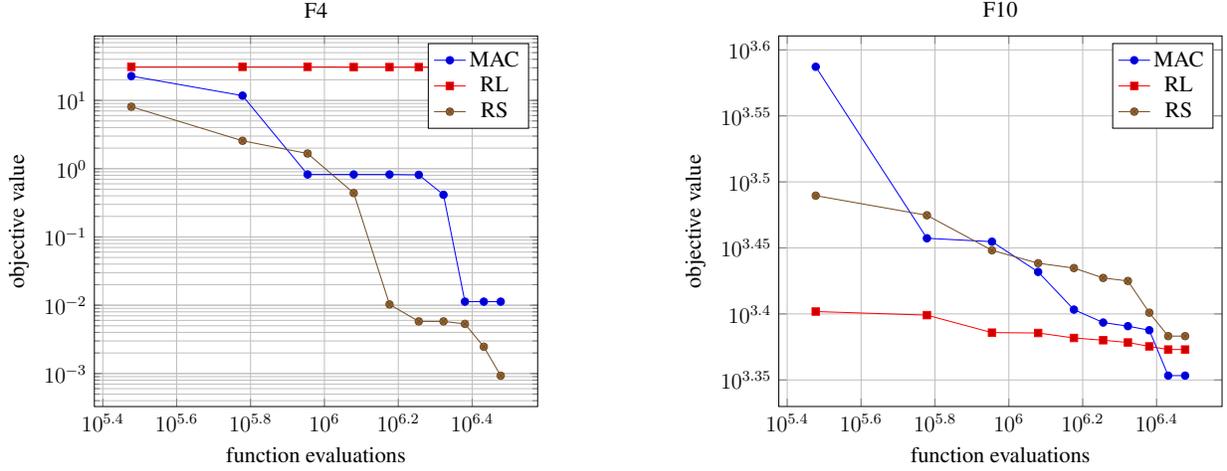
\begin{figure*}

\begin{tikzpicture}[scale=0.7]
\begin{loglogaxis}[grid=both,xlabel=function evaluations, line width=0.25, ylabel=objective value, title=F4]

\addplot coordinates {
( 300000 , 22.698853491445902 )
( 600000 , 11.731596372385216 )
( 900000 , 0.8220154880882161 )
( 1200000 , 0.8216345160882159 )
( 1500000 , 0.8216345160882159 )
( 1800000 , 0.8122510347762161 )
( 2100000 , 0.4148394272362159 )
( 2400000 , 0.011285727792516 )
( 2700000 , 0.011285727792516 )
( 3000000 , 0.011285727792516 )
};

\addplot coordinates {
(300000 , 30.9640275508708 )
(600000 , 30.9560209125558 )
(900000 , 30.949135020418804 )
(1200000 , 30.747597510418803 )
(1500000 , 30.741329960418803 )
(1800000 , 30.741329960418803 )
(2100000 , 30.7048210104188 )
(2400000 , 30.704016110418802 )
(2700000 , 30.5046970694558 )
(3000000 , 30.5046970694558 )
};

\addplot coordinates {
( 300000 , 8.074453481719999 )
( 600000 , 2.5601526355562 )
( 900000 , 1.666522872595 )
( 1200000 , 0.43970405976105004 )
( 1500000 , 0.010284760801210002 )
( 1800000 , 0.00579763431321 )
( 2100000 , 0.00579763431321 )
( 2400000 , 0.00532845176811 )
( 2700000 , 0.00246564505811 )
( 3000000 , 0.0009298076614100001 )
};

\legend{MAC, RL, RS}
\end{loglogaxis}
\end{tikzpicture}
\hskip 50pt
\begin{tikzpicture}[scale=0.7]
\begin{loglogaxis}[grid=both,xlabel=function evaluations, line width=0.25, ylabel=objective value, title=F10]

\addplot coordinates {
( 300000 , 3867.5314449999996 )
( 600000 , 2867.5314449999996 )
( 900000 , 2850.6414059999997 )
( 1200000 , 2704.181738 )
( 1500000 , 2531.984376 )
( 1800000 , 2475.5730479999997 )
( 2100000 , 2459.935645 )
( 2400000 , 2442.6879879999997 )
( 2700000 , 2256.796289 )
( 3000000 , 2256.796289 )
};

\addplot coordinates {
( 300000 , 2523.4456940000005 )
( 600000 , 2507.7564440000006 )
( 900000 , 2432.400878 )
( 1200000 , 2430.96787 )
( 1500000 , 2409.853417 )
( 1800000 , 2400.349121 )
( 2100000 , 2391.070019 )
( 2400000 , 2374.73623 )
( 2700000 , 2361.7889640000003 )
( 3000000 , 2361.7889640000003 )
};

\addplot coordinates {
( 300000 , 3088.7653330000003 )
( 600000 , 2984.9613300000005 )
( 900000 , 2808.151077 )
( 1200000 , 2745.4908220000007 )
( 1500000 , 2722.2420920000004 )
( 1800000 , 2675.8804710000004 )
( 2100000 , 2661.8232450000005 )
( 2400000 , 2518.5025410000003 )
( 2700000 , 2417.6281219999996 )
( 3000000 , 2417.6281219999996 )
};

\legend{MAC, RL, RS}
\end{loglogaxis}
\end{tikzpicture}
\caption{Illustration of the convergence results obtained for minimizing F$4$ (left) and F$10$ (right) $30$-dimensional functions.  We show the average objective value found by the competitive methods as a function of evaluations. 
}
\label{figure:conv_rate}%
\end{figure*}

\textbf{In Fig.~\ref{figure:conv_rate}}, the convergence rate of the MAC, RS, and RL search methods are shown, where the horizontal axis indicates the average objective value found by the competitive methods as a function of evaluations. We can see that MAC surpasses the performance of the RS beyond the first stages of evolution for both F$4$ and F$10$ functions. Regarding the final results, however, RS results in faster learning and eventually converges to better NAS settings for F$4$. Moreover, the advantage of RL can be seen on $F10$, although it does not perform as well on the other test functions. This undesirable performance could be alleviated if the hyperparameters of RL are chosen to be configured properly for each problem; in which case the advantage of MAC and RS parameter-free methods can be better manifested. Having this in mind, all the following experiments will be done using the MAC search strategy.

\textbf{In Table~\ref{table:nas-jso}}, the experimental results for $50$ and $100$-dimensional problems are reported. The results are evaluated against the state-of-the-art jSO algorithm \footnote{The code for jSO is publicly available at https://github.com/P-N-Suganthan/CEC2017-BoundContrained.git}. The extensions of the differential evolution are always among the winners of the CEC competition. The algorithm is shown to outperform LSHADE~\cite{tanabe2014improving} (the winner of the CEC 2014) and its new extension for CEC 2016 (iL-SHADE~\cite{brest2016shade})  which motivated us to consider jSO~\cite{7969456} algorithm for the purpose of comparison. In order to make a fair comparison, all the experiment conditions are the same as mentioned before. The obtained performance for the unimodal function F$1$ indicates that jSO gives a more accurate range for $50$ and $100$-dimensional cases, although NAS has a more reasonable performance on $100$ dimension. Conversely, it can be observed that NAS has provided better results for F$4$ test problems. Furthermore, it can be seen that jSO converges closer to global optimum for $50$ and $100$-dimensional function F$6$. Concerning the other functions, the introduced NAS methodology achieved the lowest minimum values. The results also indicate that the NAS method can be also less sensitive to the increases in dimension as well as jSO evolutionary algorithm. 


\begin{table}[htb]
      \centering
      \caption{The performance of NAS and jSO on CEC~2017 for $50$ and $100$ set cases using Wilcoxon signed-rank with $\alpha=0.05$. 
      }
        \begin{tabular}{l|l|l|l|l}
        \midrule
        \makecell {No\quad \quad \quad \quad \quad} & {NAS (50)} & {jSO (50)} & {NAS (100)}  & {jSO (100)} \\
         \midrule
         F$1$       &  $1.27460e+01$    &   $0.00000e+00$    &   ${3.63306e-03}$ &   $0.00000e+00$ \\
         F$3$       &  $1.66479e+04$    &   $0.00000e+00$    &   ${1.79154e+04}$ &   $0.00000e+00$ \\
         F$4$       &  $0.00000e+00$    &   $2.85127e+01$    &   ${5.72086e-02}$ &   $1.97357e+02$ \\
         F$5$       &  $4.63491e+00$    &   $1.19395e+01$    &   ${1.76348e+01}$ &   $2.48740e+01$ \\
         F$6$       &  $3.39670e-02$    &   $0.00000e+00$    &   ${4.73965e-02}$ &   $0.00000e+00$ \\
         F$7$       &  $4.69719e+00$    &   $6.10567e+01$    &   ${1.04077e+01}$ &   $1.28151e+02$ \\
         F$8$       &  $5.86165e+00$    &   $1.19395e+01$    &   ${9.17544e+00}$ &   $2.68639e+01$ \\
         F$9$       &  $0.00000e+00$    &   $0.00000e+00$    &   ${0.00000e+00}$ &   $0.00000e+00$ \\
         F$10$      &  $6.19778e+02$    &   $2.32882e+03$    &   ${1.34090e+03}$ &   $7.23914e+03$ \\
         \midrule
         $+ \setminus = \setminus -$   &       &   $5 \setminus 3 \setminus 1$    &   &  $5 \setminus 3 \setminus 1$ \\
        \midrule

        \end{tabular}
      \label{table:nas-jso}%
\end{table}%

Altogether, these promising results: (1) provide evidence for applying the neural networks on high dimensional optimization problems;  and (2) show that the performance of modern machine learning components for designing new optimization algorithms can be highly significant. For example, the obtained results on F$7$, F$9$, and F$10$  functions are reported  for the first time in this study. This is quite interesting because the NAS doesn't borrow any search components from  the previously proposed methods for the CEC problems.

\begin{figure*}[htb]
\centering
\begin{subfigure}{.52\textwidth}
  \centering
  \includegraphics[width=\linewidth]{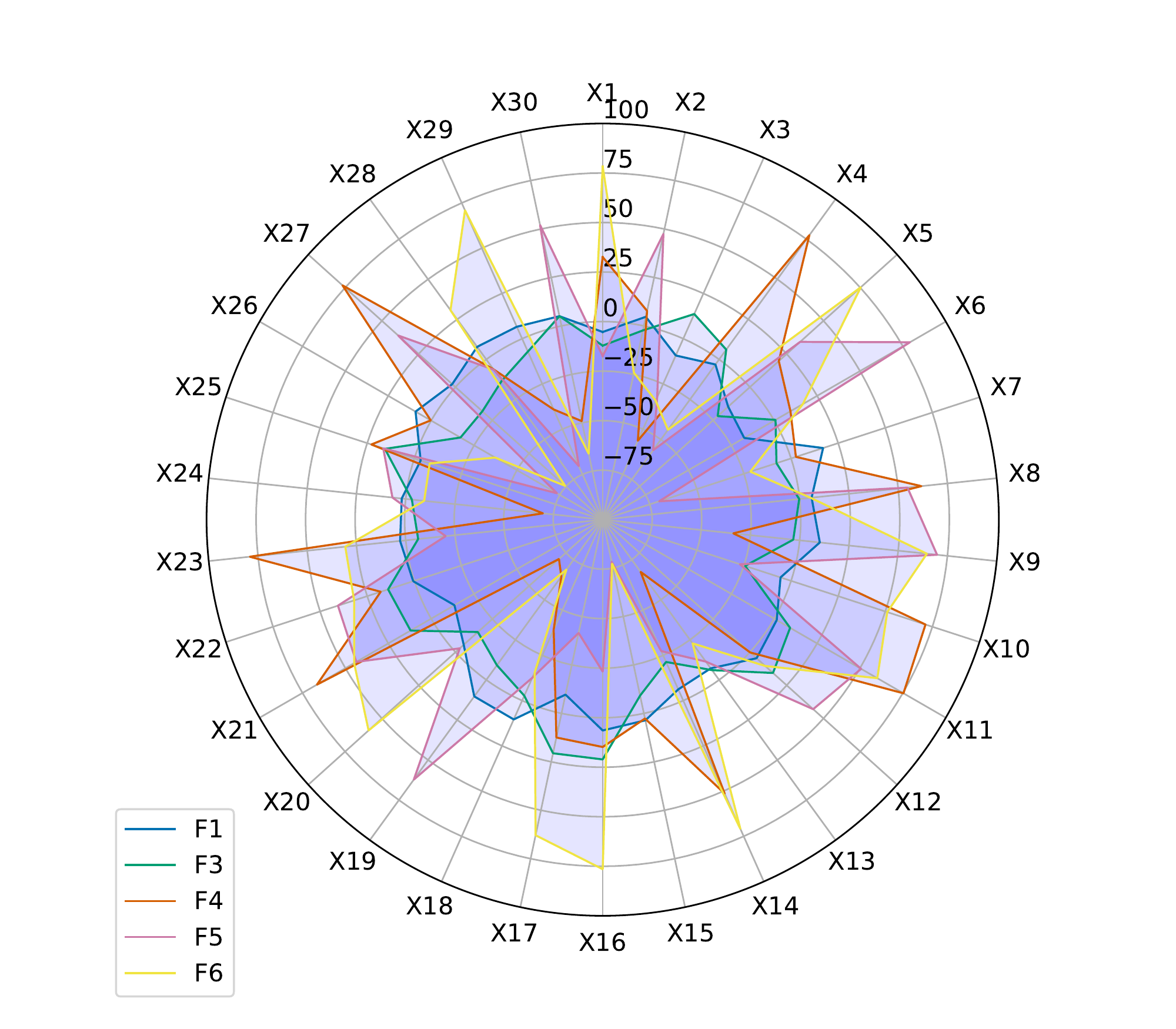}

  \label{fig:sfig1}
\end{subfigure}%
\begin{subfigure}{.52\textwidth}
  \centering
  \includegraphics[width=\linewidth]{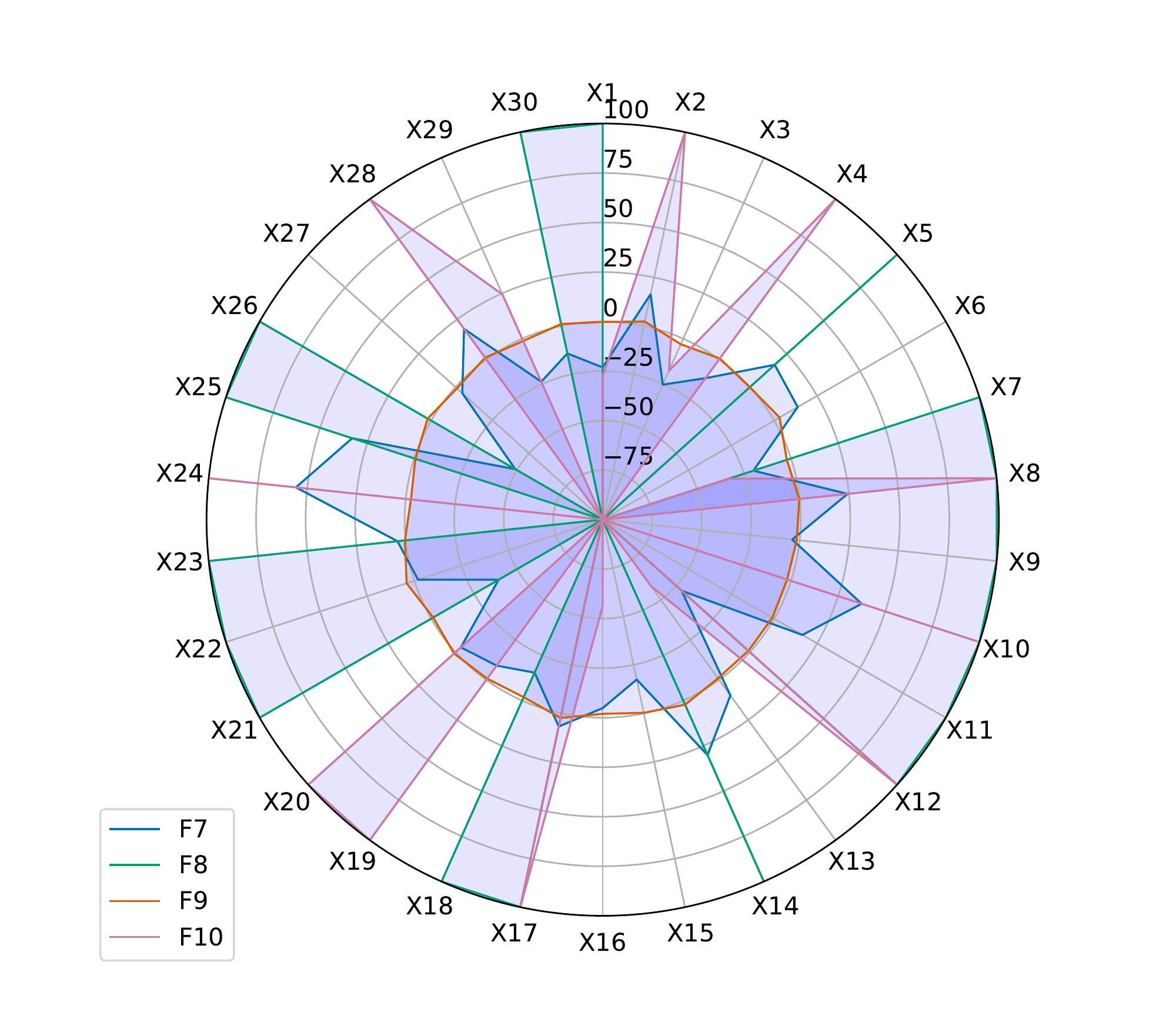}

  \label{fig:sfig2}
\end{subfigure}

\caption{Radar chart comparing the variation of the best-obtained solutions by NAS on F$1$, F$3$-F$10$ for $30$- dimensional problems. Each axis represents a quantity for a different dimension. This chart shows how the designed neural architectures for different problems found the solutions that are rather different or disparate from each other in the search space.}
\label{fig:radar-chart}
\end{figure*}

\textbf{In Fig.~\ref{fig:radar-chart}}, the radar chart of the obtained solutions using NAS for 30-dimensional data in the form of a plot is presented. We can see that the best-found solutions for the CEC problems are rather different or disparate from each other in the search space. However, it can be seen that the generated neural networks are able to learn the correlation between the decision variables. Accordingly, we can say that the introduced NAS methodology is not a simple biased search method that generates solutions only around a specific area in the search space.

\textbf{In Tables~\ref{table:tf-50d} and~\ref{table:tf-100d}}, we presented the obtained results by transferring information from previous runs on the higher dimension problems. We show how the trained neural architectures for $30$-dimensional problems can generalize well on higher dimensions $50$ and $100$. Our task is to take the best pre-trained architecture for each problem and transfer its weights to a higher dimension problem; rather than searching and training a model from scratch. In the following experiment, the batch size will default to $1$. To have a fair comparison, five different scenarios are considered in this study. In NAS-$1$, the weights of the trained layers of the network are frozen to use previously learned weights that are hidden throughout all the layers. Thereafter, we change and re-train the last fully-connected Dense layer according to the new dimension. For NAS-$2$, however, all the weights are re-initialized randomly and architecture should learn everything from scratch. In jSO-$2$, the solution vectors are initialized with the best-found solution for the $30$ dimensions, while the algorithm is still able to change the initialized parameters. Alternatively, we freeze these parameters in jSO-$3$, which reduces the dimensionality of the problem and helps the algorithm to have a better search efficiency~\footnote{Note that the jSO has a population reduction schema and it is not possible to transfer the final population to a new optimization task. More importantly, we notice that the population converges to the best-found solution for all the CEC problems.}. We would like to explore the possibility of speeding up the search process so as to avoid the computational overhead of re-optimizing for large scale computationally expensive problems. To  answer this,  we repeat the experiments with a cutoff of $1000$ evaluations over $15$ different runs. In Table~\ref{table:tf-50d}, we can see that the proposed NAS-$1$ significantly outperformed the randomly initialized architectures NAS-$2$ and also all the extensions of the jSO; by switching  the  roles  of  the  classical evolutionary algorithms and learning distributions of the good solutions. The results for NAS-$1$ and NAS-$2$ suggest that not only the architecture, but also the learned weights of the network are crucial to the success of accelerating the optimization process.  Notably, NAS-$1$ yielded more scalable performance for the $100$-dimensional problems. Overall, we can say the neural architectures addressed the practical limitation of
learning from previous similar problems with different dimensions; compared to evolutionary and surrogate-assisted algorithms. Under this setting, we will be able to transfer the parameters of a neural network from a cheap-to-evaluate problem to another similar one in order to accelerate the search process. We will show that these results can similarly generalized to the protein problems.
\begin{table}[htb]
      \centering
      \caption{The performance of NAS and jSO on $50$-dimensional CEC~$2017$ test set. The results are obtained by exploiting the performance of NAS and jSO on $30$-dimensional benchmarks.
      }
        \resizebox{0.95\textwidth}{!}{%
        \begin{tabular}{l|l|l|l|l|l}
        \midrule
        \makecell {No\quad \quad \quad \quad \quad} & {NAS-1} & {NAS-2} & {jSO} & {jSO-2} & {jSO-3}\\
         \midrule
             F1       &  $2.96902e+03$     &   $2.41835e+11$    &   $1.62804e+11$  &   $4.66952e+10$ & $1.734390e+10$\\
             F3       &  $1.81075e+05$     &   $2.31981e+05$    &   $2.35311e+05$  &   $8.10353e+04$ & $7.270594e+04$\\
             F4       &  $1.48287e+02$     &   $4.51499e+02$    &   $4.74709e+04$  &   $6.82198e+03$ & $2.358366e+03$\\
             F5       &  $2.90176e+02$     &   $1.95285e+03$    &   $9.00615e+02$  &   $5.13882e+02$ & $4.671343e+02$\\
             F6       &  $6.30965e+01$     &   $1.40318e+02$    &   $1.25437e+02$  &   $1.32398e+01$ & $8.421186e+00$\\
             F7       &  $8.26953e+02$     &   $6.28353e+03$    &   $3.49137e+03$  &   $9.54540e+02$ & $8.665667e+02$\\
             F8       &  $4.08987e+02$     &   $1.21316e+03$    &   $8.77639e+02$  &   $5.41581e+02$ & $3.926943e+02$\\
             F9       &  $1.14396e+04$     &   $6.78905e+04$    &   $6.81788e+04$  &   $1.34382e+04$ & $4.873473e+03$\\
             F10      &  $6.74823e+03$     &   $1.43720e+04$    &   $1.43100e+04$  &   $1.46106e+04$ & $1.470171e+04$\\
        \midrule
         $+ \setminus = \setminus -$   &       &   $9 \setminus 0 \setminus 0$    &  $9 \setminus 0 \setminus 0$ & $7 \setminus 2 \setminus 0$ & $6 \setminus 3 \setminus 0$\\
        \midrule

        \end{tabular}}
      \label{table:tf-50d}%
\end{table}%

\begin{table}[htb]
      \centering
      \caption{The performance of NAS and jSO on $100$-dimensional CEC~$2017$ test set. The results are obtained by exploiting the performance of NAS and jSO on $30$-dimensional benchmarks.
      }
        \resizebox{0.95\textwidth}{!}{%
        \begin{tabular}{l|l|l|l|l|l}
        \midrule
        \makecell {No\quad \quad \quad \quad \quad} & {NAS-1} & {NAS-2} & {jSO} & {jSO-2} & {jSO-3}\\
         \midrule
             F1       &  $1.28832e+04$     &   $5.47524e+11$    &   $4.99146e+11$  &   $2.62497e+11$ &   $2.85013e+11$ \\
             F3       &  $3.49607e+05$     &   $5.29566e+05$    &   $7.70880e+05$  &   $5.80964e+05$ &   $5.45895e+05$\\
             F4       &  $1.71629e+02$     &   $5.51000e+02$    &   $1.77581e+05$  &   $8.21492e+04$ &   $5.60870e+04$\\
             F5       &  $6.55163e+02$     &   $4.59244e+03$    &   $2.27627e+03$  &   $1.72975e+03$ &   $1.69408e+03$\\
             F6       &  $6.10705e+01$     &   $1.60571e+02$    &   $1.52327e+02$  &   $6.19906e+01$ &   $6.60414e+01$\\
             F7       &  $1.36100e+03$     &   $1.52395e+04$    &   $1.10122e+04$  &   $7.34962e+03$ &   $7.27311e+03$\\
             F8       &  $1.07242e+03$     &   $2.80098e+03$    &   $2.24957e+03$  &   $1.88354e+03$ &   $1.88620e+03$\\
             F9       &  $2.23360e+04$     &   $1.34184e+05$    &   $1.79658e+05$  &   $1.13661e+05$ &   $9.97511e+04$\\
             F10      &  $1.46885e+04$     &   $2.76060e+04$    &   $3.23123e+04$  &   $3.25155e+04$ &   $3.16535e+04$\\
        \midrule
         $+ \setminus = \setminus -$   &       &   $9 \setminus 0 \setminus 0$    &  $9 \setminus 0 \setminus 0$ &  $9 \setminus 0 \setminus 0$ &  $9 \setminus 0 \setminus 0$\\
        \midrule

        \end{tabular}}
      \label{table:tf-100d}%
\end{table}%

\begin{figure*}[!ht]

\begin{subfigure}{.33\textwidth}

  \includegraphics[width=\linewidth]{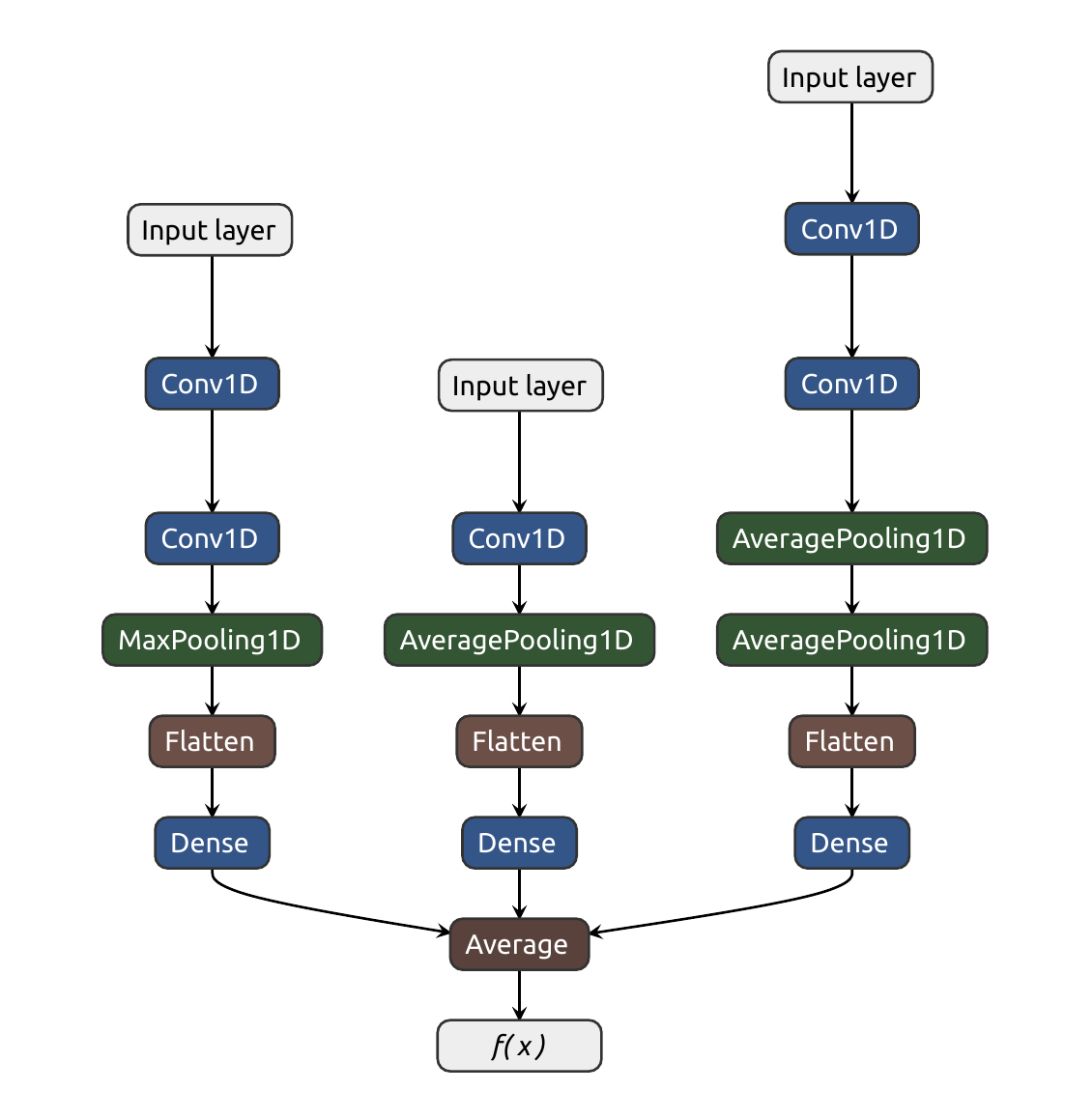}
  \caption{Bagging}
  \label{fig:ensemble-plot:1}
\end{subfigure}%
\begin{subfigure}{.33\textwidth}

  \includegraphics[width=\linewidth]{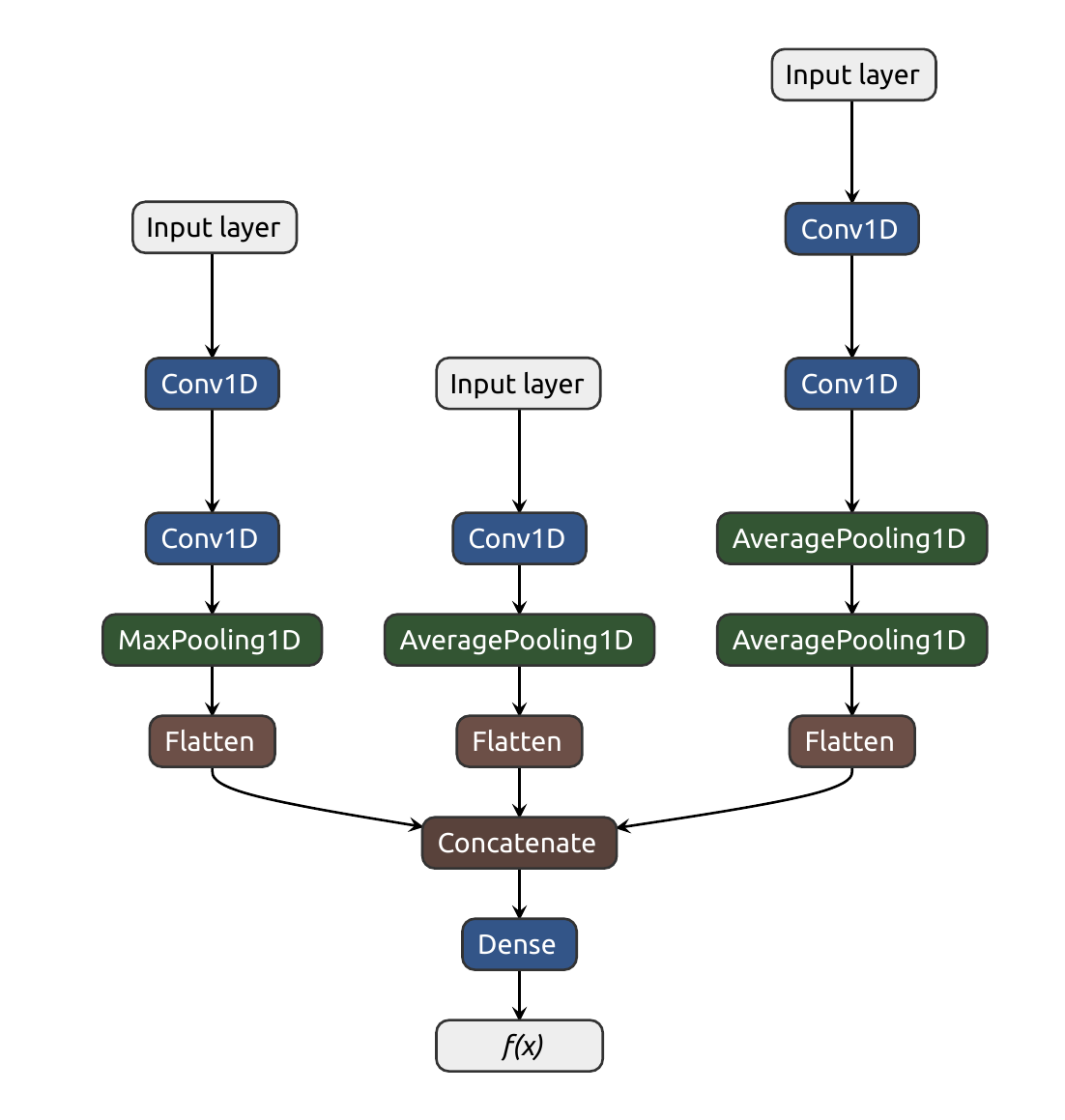}
  \caption{Stacking}
  \label{fig:ensemble-plot:2}
\end{subfigure}
\begin{subfigure}{.33\textwidth}

  \includegraphics[width=\linewidth]{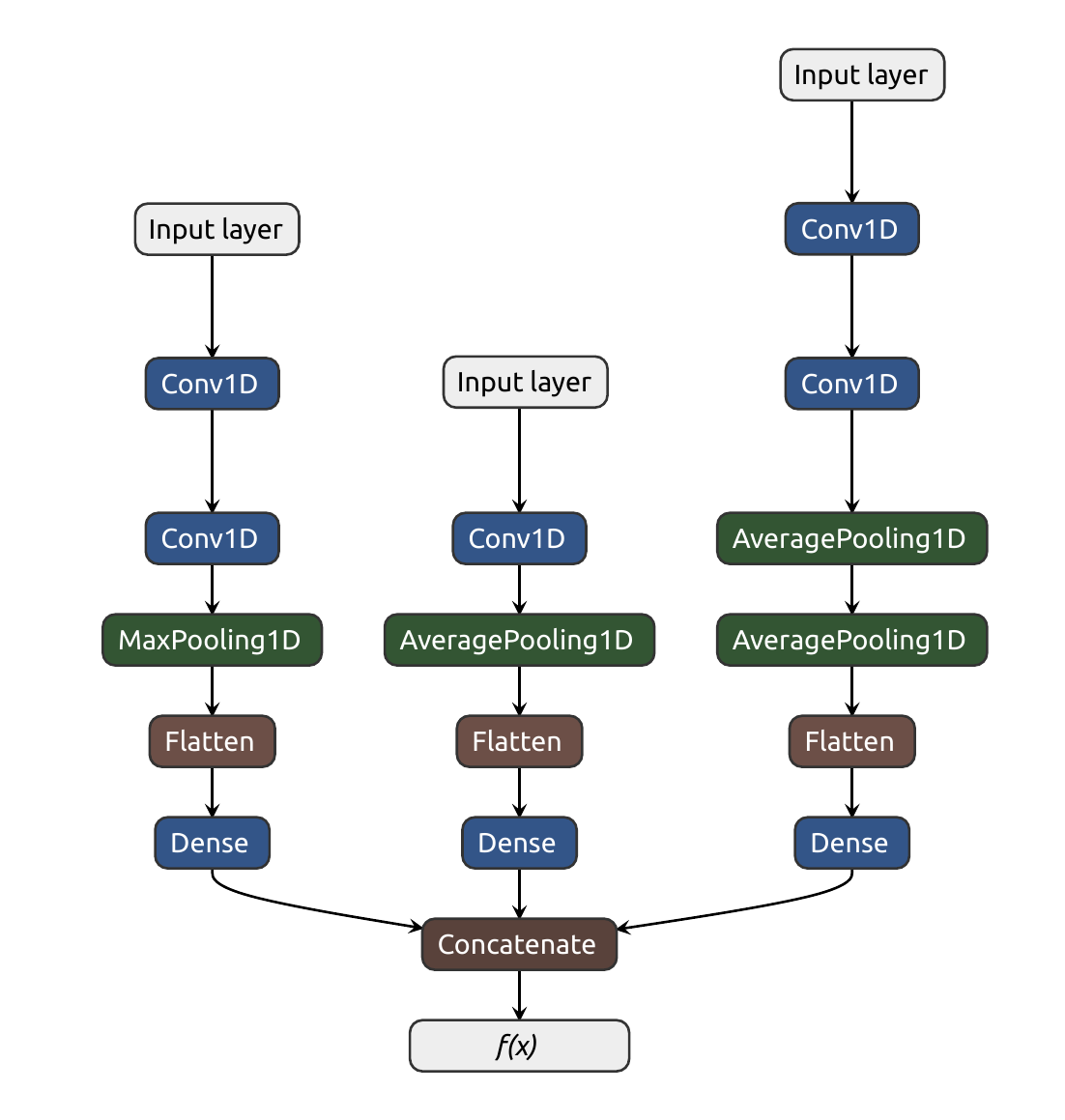}
  \caption{Hybrid}
  \label{fig:ensemble-plot:3}
\end{subfigure}

\caption{Easy visualization of different ensemble learning methods using weak learners. In this particular case, the results of the models are aggregated to yield a better performance. The Bagging ensemble model is constructed by averaging the final results from each learner, while the idea of the Stacking model is to mix different weak learners by training a meta-model. Besides these two, Hybrid model is a new approach we introduced for optimization tasks.}
\label{fig:ensemble-plot}
\end{figure*}

\textbf{In Fig.~\ref{fig:ensemble-plot}}, three different ensemble learning strategies are illustrated\footnote{Netron Visualizer is used to illustrate the model. The tool is available online at \url{https://github.com/lutzroeder/netron}}. Accordingly, we conduct experiments to show the performance of multiple models instead of a single one for obtaining the best possible results to solve  the $50$ and $100$  problems  under a limited budget. In machine learning, this is called ensemble learning and tends to yield better  performance by averaging the results over multiple models. In Fig.~\ref{fig:ensemble-plot:1}, the main principle is to fit a set of weak learners in parallel and to combine them following deterministic averaging. The term “Bagging” is used to describe a family of such ensemble methods~\cite{dietterich2000ensemble}. Fig.~\ref{fig:ensemble-plot:2} illustrates the ensemble Stacking model where the extracted information from the models is combined. This blending process can be defined using one or more additional layers. In this study, we only going to use one Dense layer for simplicity. Fig.~\ref{fig:ensemble-plot:3} shows a new hybrid ensemble method that we proposed specifically for optimization tasks. In this strategy, we build and train an ensemble model based on the average value of the outputs (i.e., objective function values) in homogeneous learners. 

The already elaborated methods are used to investigate whether we can improve the results in Tables~\ref{table:tf-50d} and~\ref{table:tf-100d}. The main hypothesis is to combine the weak learners so as to improve the results returned by the base models. The ensemble learning is applied to the obtained models for $30$-dimensional problems for solving $50$ and $100$ dimensions. Similar to the previous experiment, the batch size will default to $1$ for all the weak learners. We applied the Bagging, Stacking, and Hybrid schemes on the pre-trained  architectures for $30$-dimensional problems over $15$ runs;  rather than  using  and  re-training  only the best model. The weights of the trained layers for all the models are frozen. Moreover, we change and re-train the last fully-connected Dense layer according to the new dimension. For the Bagging and Stacking methods, we
repeat the experiments with a cutoff of 1000 evaluations over $15$  different  runs, while the Hybrid model needs $15 \times 1000$ evaluations.

\textbf{From Tables~\ref{table:tf-50d} and~\ref{table:ensemble-50d}} we can see that the Bagging method outperforms the NAS-$1$ on functions F$1$, F$4$, and F$6$-F$8$. Furthermore, the Stacking model gives better results on F$4$, F$7$-F$8$, and F$10$. Needless to say, the results illustrate  the  significant  improvement  of  the Hybrid model.

\begin{table}[htb]
      \centering
      \caption{Performance of NAS on $50$-dimensional  CEC~$2017$ test set using ensemble learning. The results are obtained by aggregating the trained models over $15$ runs on $30$-dimensional benchmarks.
      }

        \begin{tabular}{l|l|l|l}
        \midrule
        \makecell {No\quad \quad \quad \quad \quad} & {Bagging} & {Stacking} & {Hybrid}\\
         \midrule
             F$1$       &  $1.63821e+03$     &   $4.78148e+03$    &   $4.37989e+02$\\
             F$3$       &  $1.11977e+06$     &   $1.42817e+06$    &   $1.66900e+05$\\
             F$4$       &  $7.74409e+01$     &   $1.09783e+02$    &   $3.35049e+01$\\
             F$5$       &  $3.14418e+02$     &   $5.67149e+02$    &   $1.01177e+02$\\
             F$6$       &  $6.19302e+01$     &   $1.26289e+02$    &   $4.60574e+01$\\
             F$7$       &  $2.97057e+02$     &   $4.99652e+02$    &   $2.63861e+02$\\
             F$8$       &  $1.86261e+02$     &   $2.37593e+02$    &   $1.52186e+02$\\
             F$9$       &  $1.22938e+04$     &   $1.31054e+04$    &   $1.06279e+04$\\
             F$10$      &  $7.02342e+03$     &  $5.72794e+03$    &  $4.99745e+03$\\
        \midrule
        \end{tabular}
      \label{table:ensemble-50d}%
\end{table}%

\textbf{In Tables~\ref{table:tf-100d} and~\ref{table:ensemble-100d}}, similar results for $100$ dimensions are recorded which offer the flexibility of the ensemble methods in proportion to the new higher scale. Overall, the obtained results demonstrate the usefulness of the three ensemble methods in the context of optimization. 

\begin{figure*}[!htbp]

\begin{subfigure}{.52\textwidth}
  \centering
  \includegraphics[width=\linewidth]{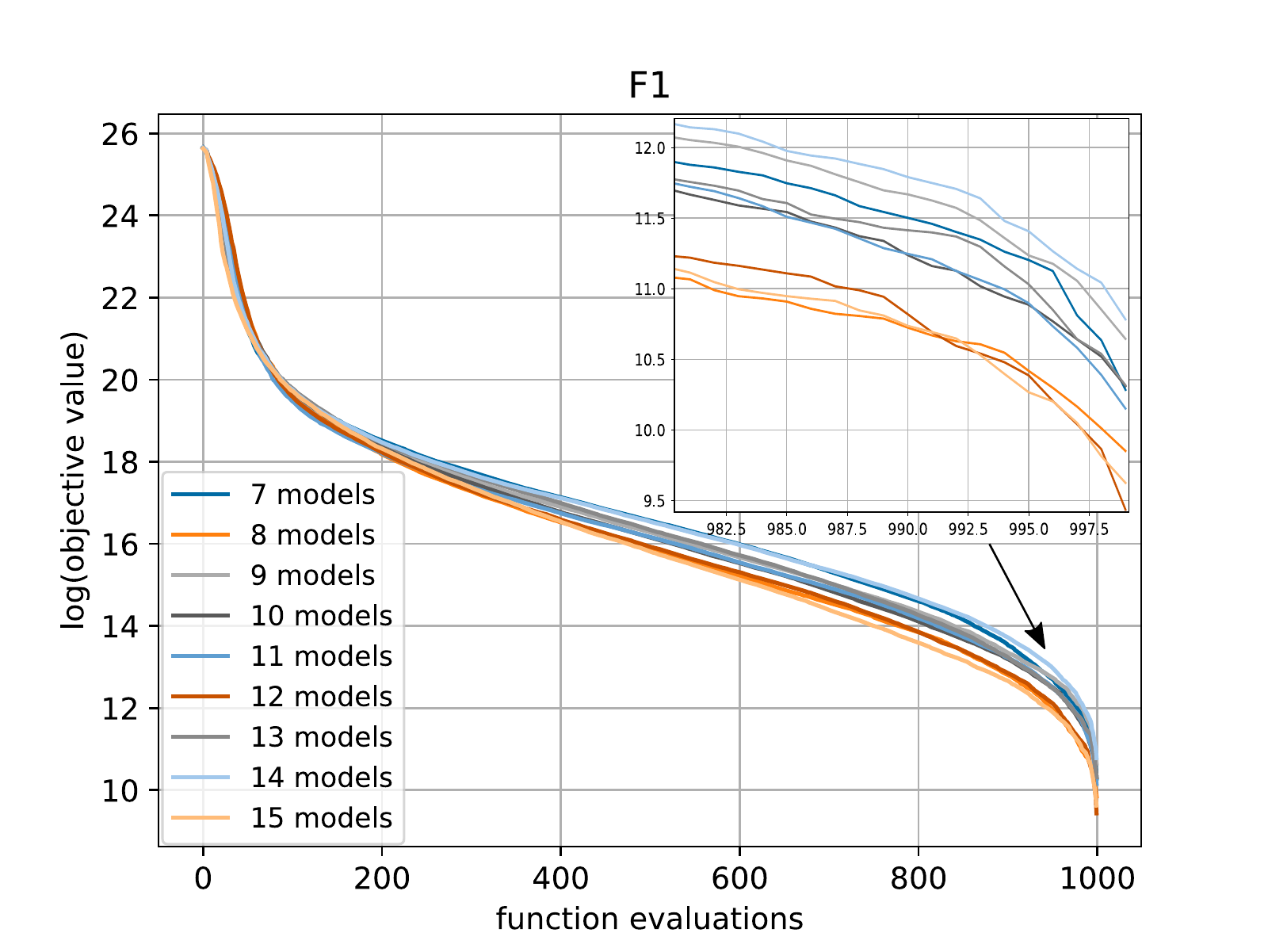}
  \caption{}
\end{subfigure}%
\begin{subfigure}{.52\textwidth}
  \centering
  \includegraphics[width=\linewidth]{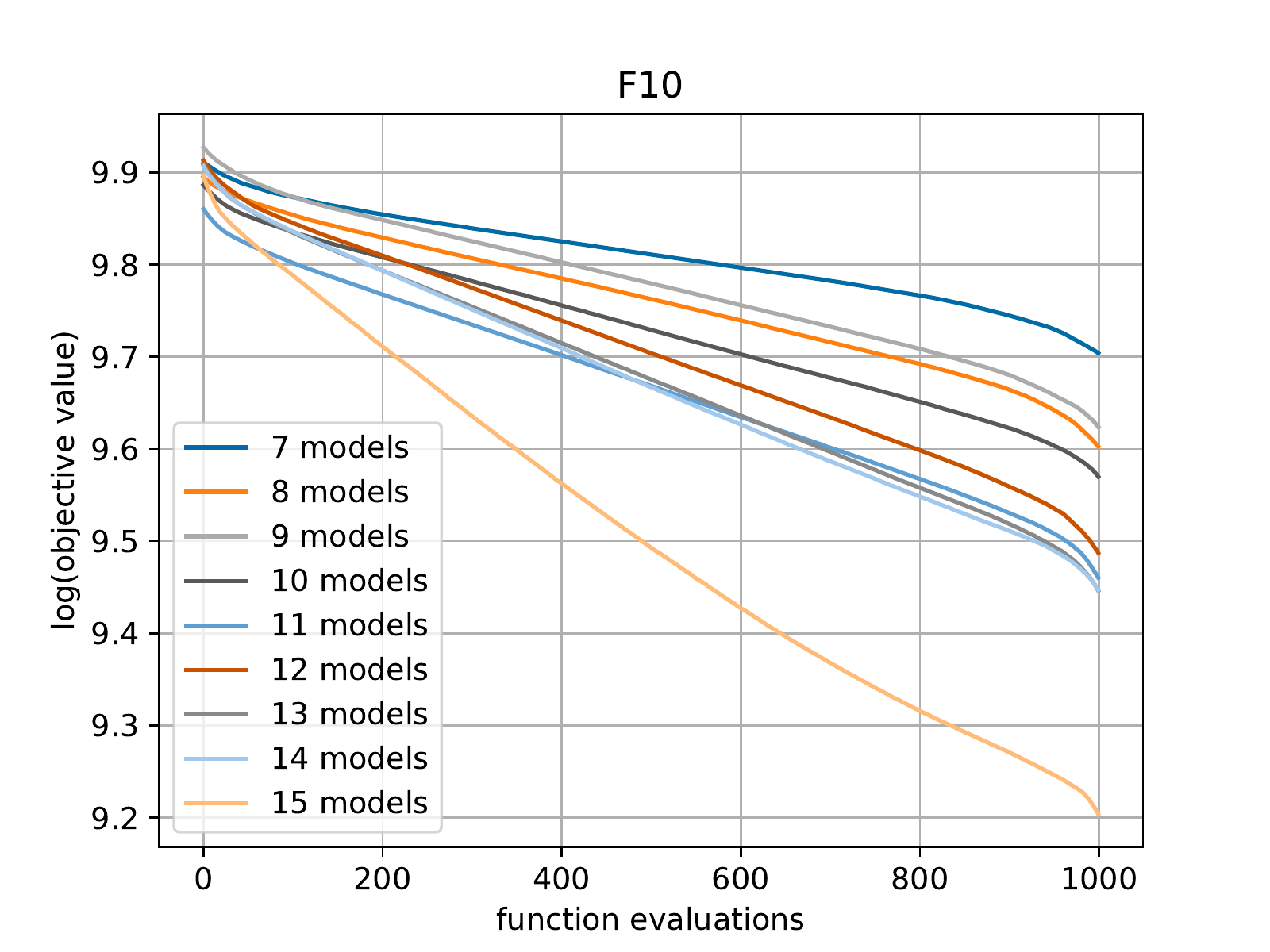}
  \caption{}
\end{subfigure}

\caption{This figure illustrates and compares the performance of the (a) Bagging and (b) Stacking ensemble methods using different numbers of models. In most cases, we can say that the larger is the number of models, the more enhanced are the results.}
\label{fig:ensemble-4}
\end{figure*}

\textbf{Fig.~\ref{fig:ensemble-4}} gives an intuitive understanding of how the number of models for the Bagging and Stacking methods is related to the accuracy of the generated ensemble model. We can see that there is an improvement in accuracy when the number of employed models increases. However, sometimes the performance drops to a lower value compared to the one with fewer models.  This is a very important factor to be kept in mind that there is no evidence of having more models necessary means a higher ensemble performance.

\begin{table}[htb]
      \centering
      \caption{Performance of NAS on $100$-dimensional  CEC~$2017$ test set using ensemble learning. The results are obtained by aggregating the trained models over $15$ runs on $30$-dimensional benchmarks.}
        \begin{tabular}{l|l|l|l}
        \midrule
        \makecell {No\quad \quad \quad \quad \quad} & {Bagging} & {Stacking} & {Hybrid}\\
         \midrule
             F$1$       &  $5.59750e+03$     &   $6.52806e+03$    &   $1.52856e+02$\\
             F$3$       &  $2.76794e+10$     &   $1.09759e+11$    &   $3.46635e+05$\\
             F$4$       &  $1.71032e+02$     &   $1.89967e+02$    &   $1.69293e+02$\\
             F$5$       &  $3.00000e+02$     &   $9.05549e+02$    &   $2.05242e+02$\\
             F$6$       &  $6.04775e+01$     &   $1.25322e+02$    &   $4.90530e+01$\\
             F$7$       &  $2.30870e+03$     &   $1.21078e+03$    &   $6.59674e+02$\\
             F$8$       &  $9.06463e+02$     &   $3.11334e+02$    &   $4.35899e+02$\\
             F$9$       &  $2.08151e+04$     &   $4.37464e+04$    &   $1.82395e+04$\\
             F$10$      &  $1.51215e+04$     &   $1.27997e+04$    &   $1.19021e+04$\\
        \midrule

        \end{tabular}
      \label{table:ensemble-100d}%
\end{table}%

\subsection{Results on protein structure prediction}
\textbf{In Table~\ref{table:proteins-res}}, we shift our focus to investigate the performance of NAS for PSP. All the experiments are repeated for 30 independent runs using the AB off-lattice model. The human-designed IFABC~\cite{LI201470}, LSHADE~\cite{tanabe2014improving}, and SGDE~\cite{RAKHSHANI2019100493} algorithms for protein structure optimization are considered. The best, worst, mean, and standard deviation of the results are reported. To provide a fair comparison, the parameters of the competitive methods are set  according to the original works. Interestingly, NAS achieves superior results compared to state-of-the-art algorithms designed by human experts. These results show the versatility and robustness of the search-generated neural architectures. Overall, Table~\ref{table:proteins-res} suggests that NAS strikes a better trade-off in every metric.

\begin{figure*}[!htb]

\begin{subfigure}{.5\textwidth}
  \centering
  \includegraphics[width=\linewidth]{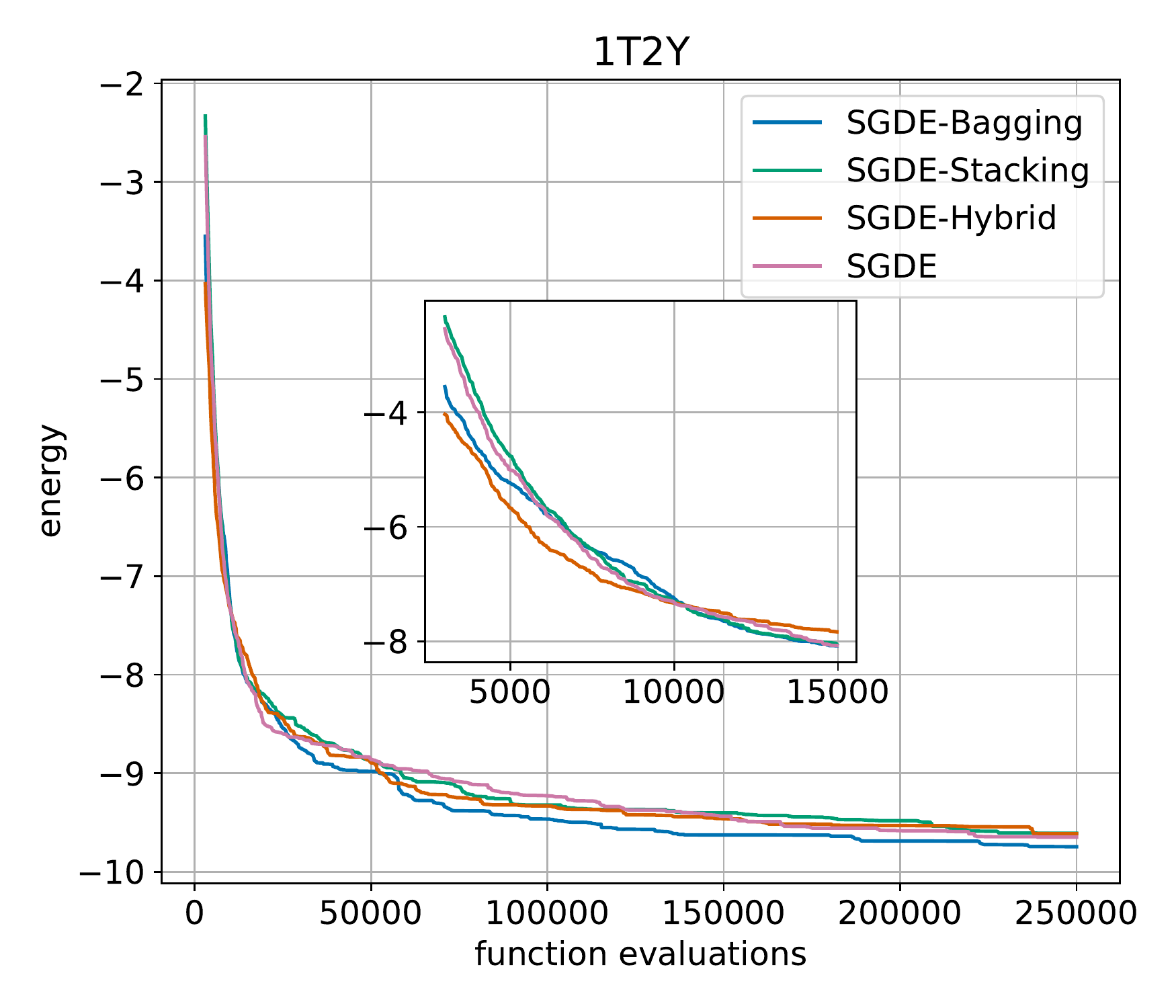}

\end{subfigure}%
\begin{subfigure}{.5\textwidth}
  \centering
  \includegraphics[width=\linewidth]{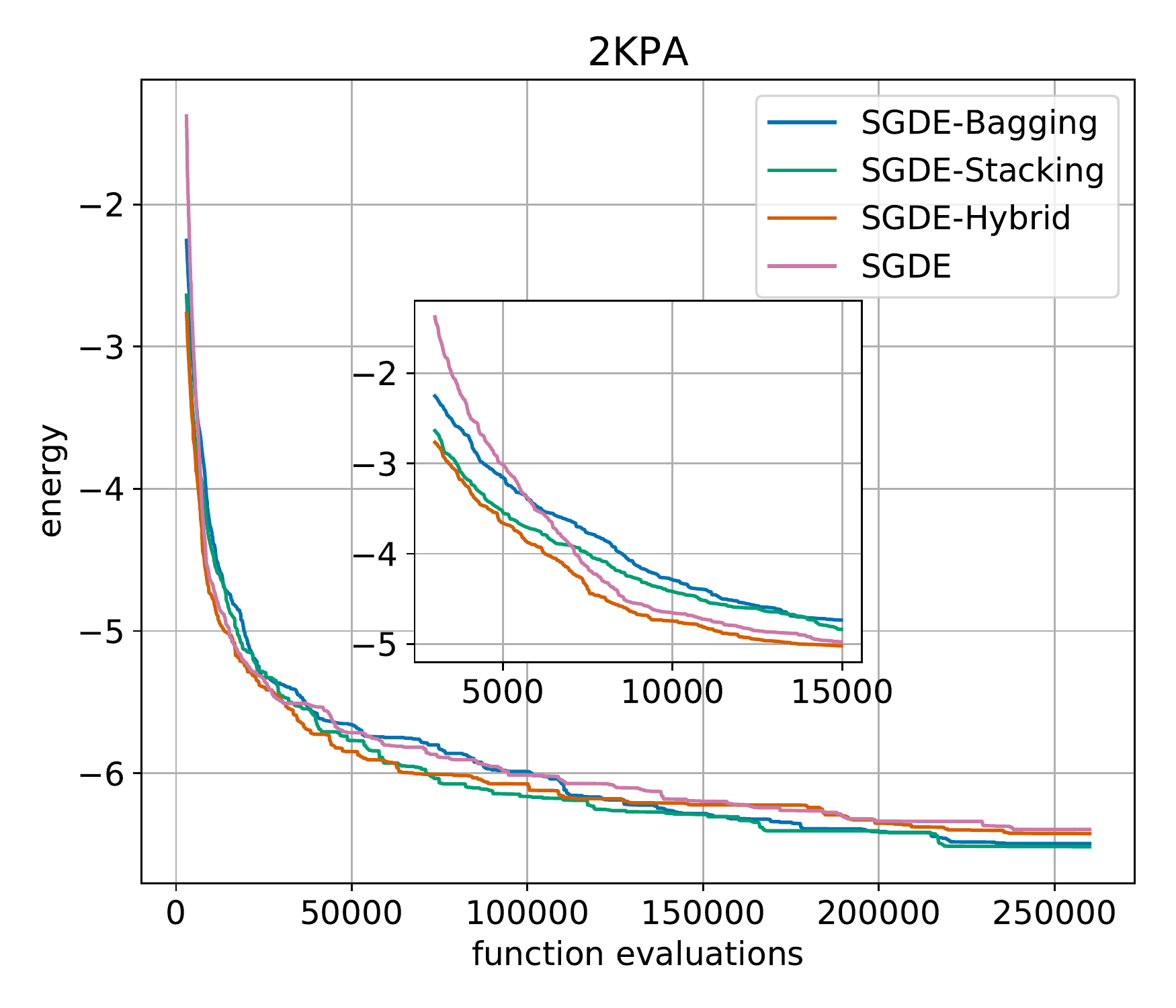}

\end{subfigure}

\begin{subfigure}{.5\textwidth}
  \centering
  \includegraphics[width=\linewidth]{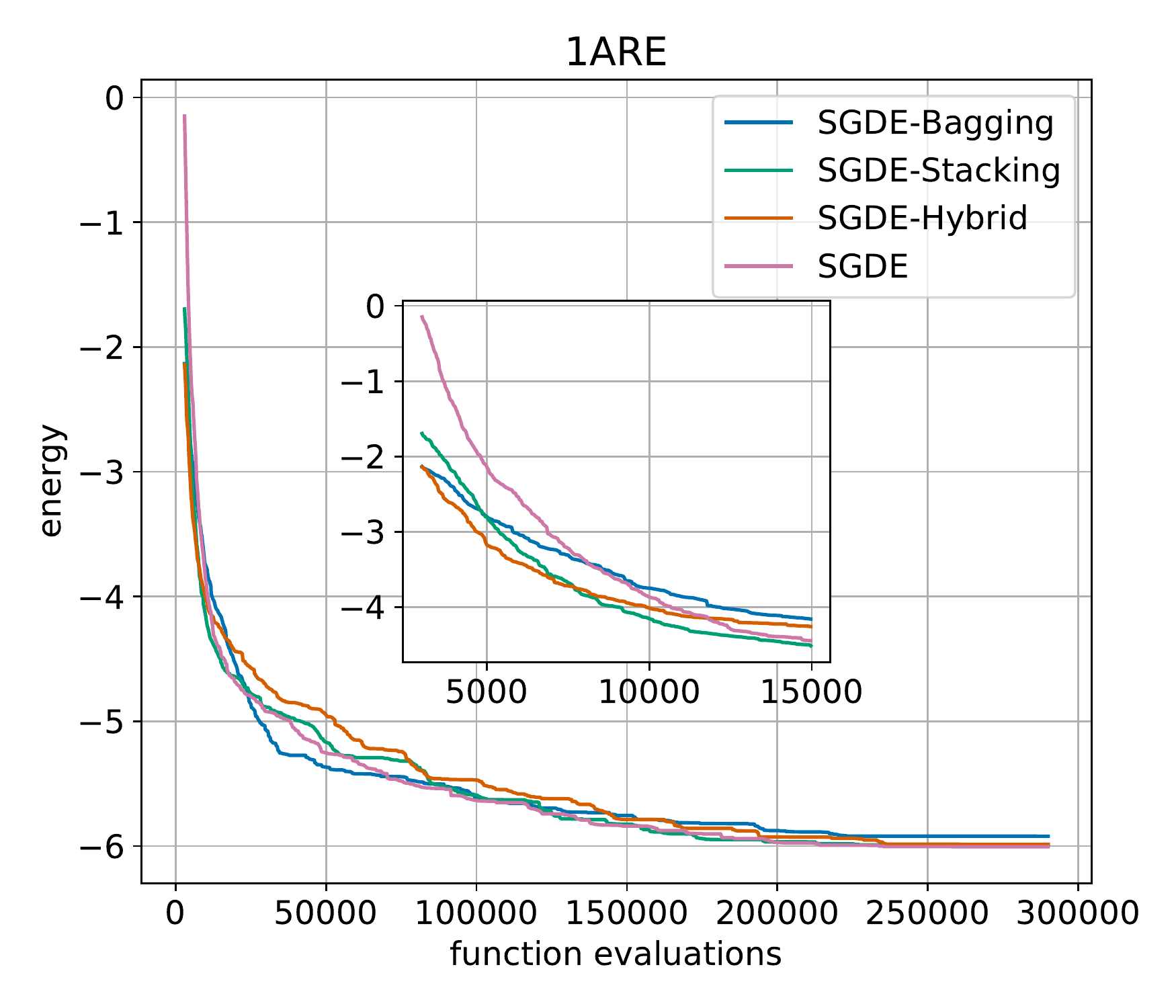}

\end{subfigure}%
\begin{subfigure}{.5\textwidth}
  \centering
  \includegraphics[width=\linewidth]{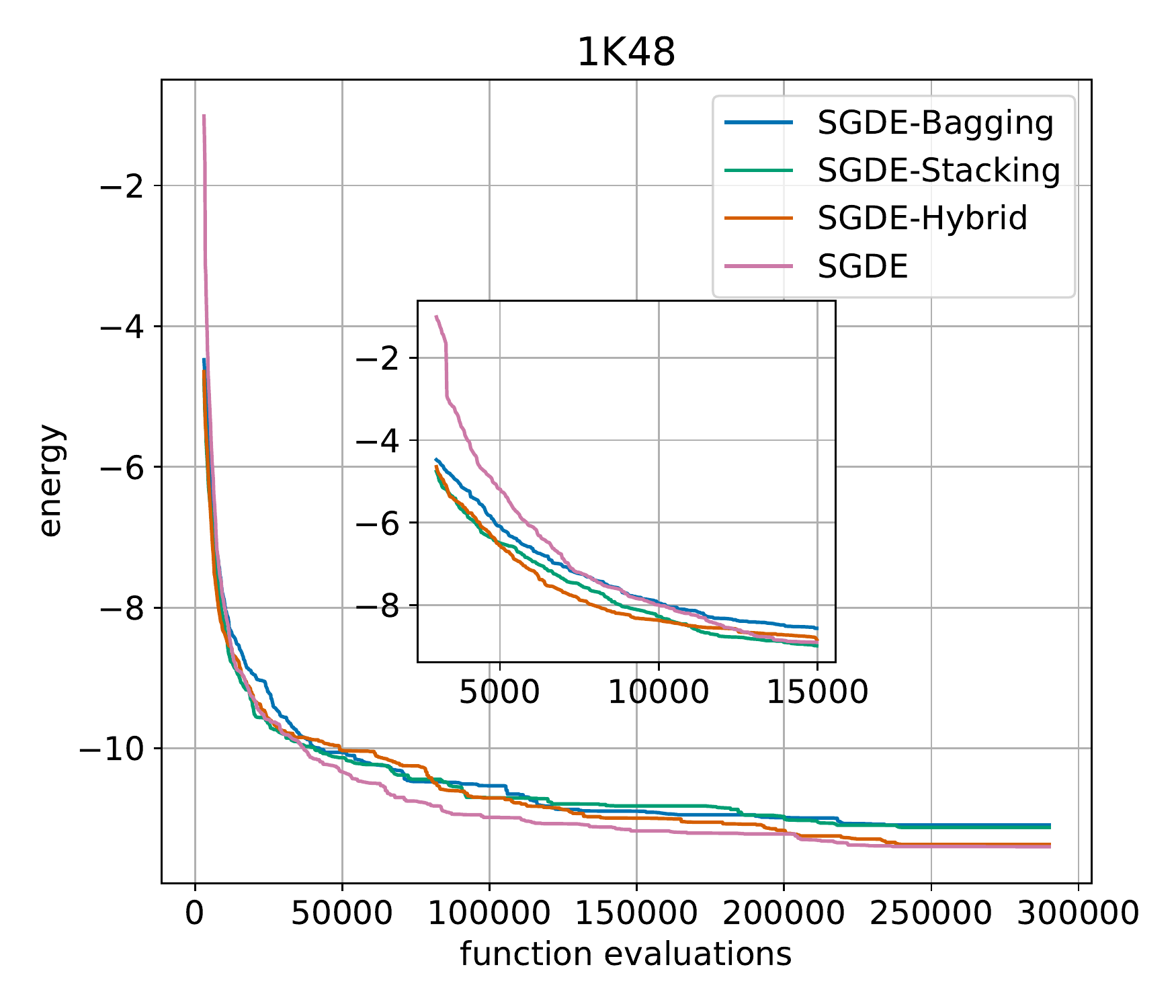}

\end{subfigure}

\begin{subfigure}{.5\textwidth}
  \centering
  \includegraphics[width=\linewidth]{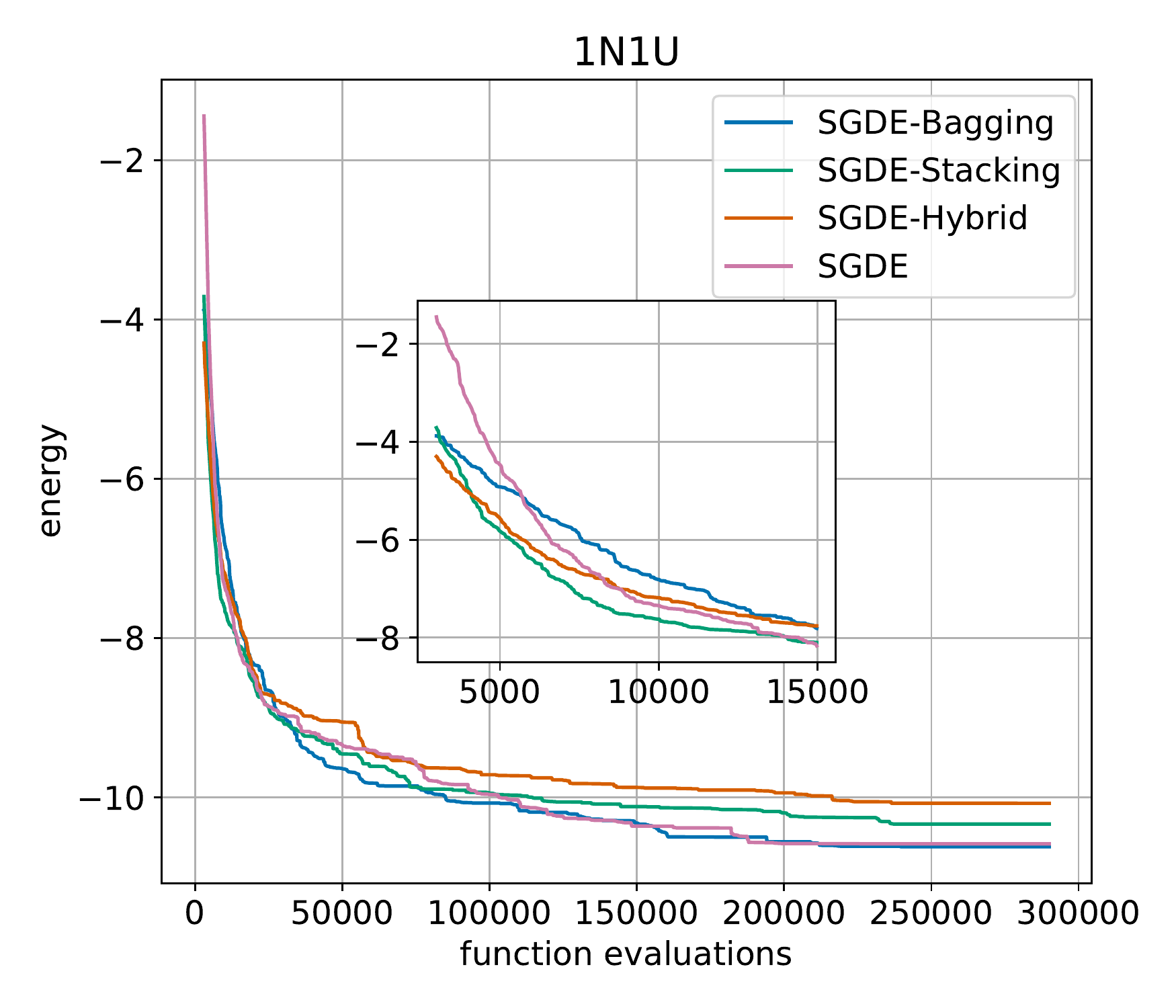}

\end{subfigure}%
\begin{subfigure}{.5\textwidth}
  \centering
  \includegraphics[width=\linewidth]{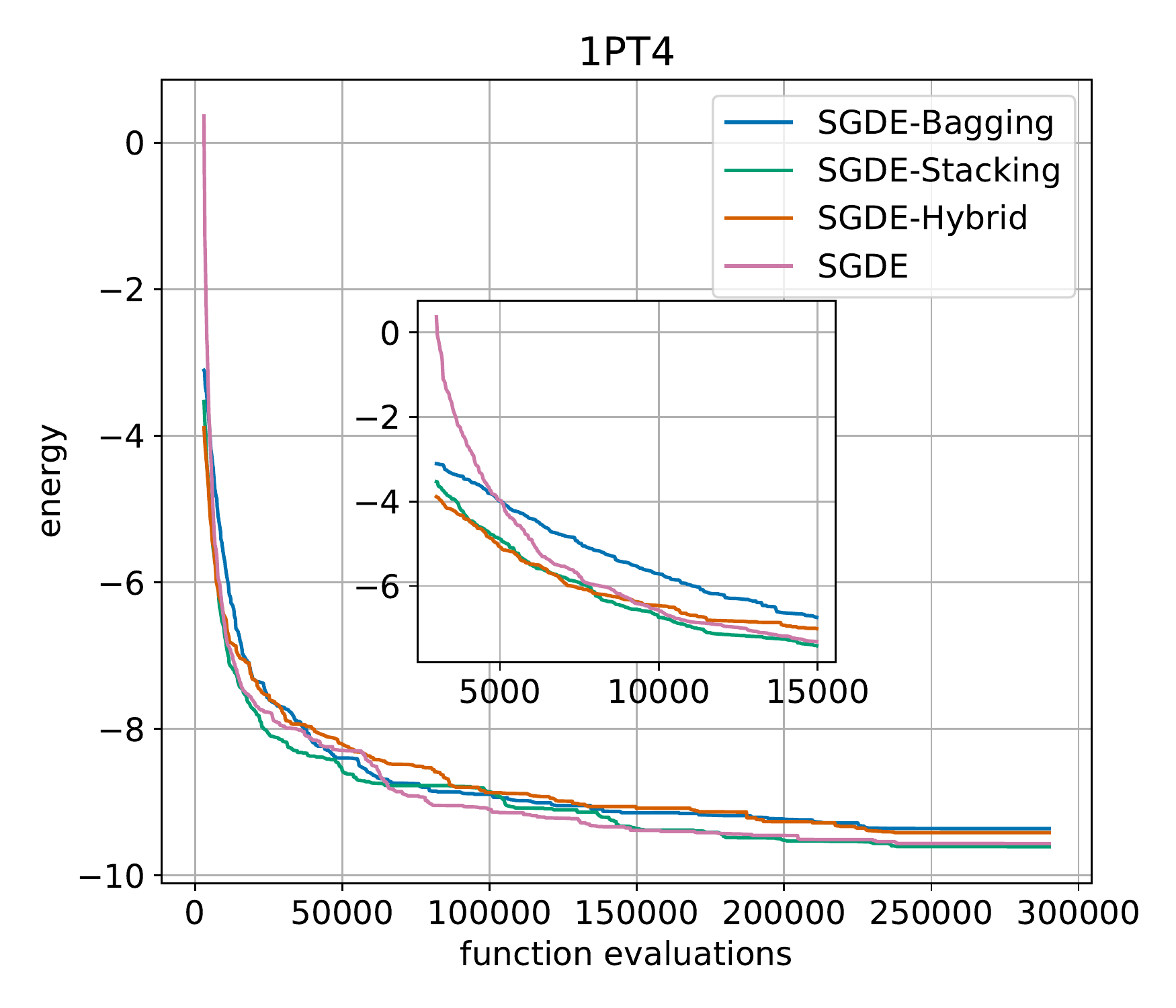}

\end{subfigure}

\caption{This figure illustrates the convergence results of SGDE obtained for $1$T$2$Y, $2$KPA, $1$ARE, $1$K$48$, $1$N$1$U, and $1$PT$4$ protein sequences with and without warm-start population initialization. We can see that the transferred information using pre-trained ensemble models enhanced the convergence performance of SGDE.
}
\label{fig:ensemble-protein-plot}
\end{figure*}

\textbf{In Fig.~\ref{fig:ensemble-protein-plot}}, we demonstrate the usefulness of NAS for evolutionary algorithms. Accordingly, the best-found architectures for solving the protein sequences $1-10$ are used to accelerate the convergence rate of SGDE on unseen protein sequences $11-16$. To do so, the ensemble model of the best architectures is first adopted to the dimensionality of the new problem; as we described before in the previous subsection. Then, we re-trained the ensemble model to generate the initial solutions for the SGDE population. The limited budget with  a  cutoff  of $500$, $500$, and $5000$ function evaluations are used for Bagging, Stacking, and Hybrid ensemble models, respectively. In SGDE, the experiments are conducted with a limited budget of $10,000 \times D$ function evaluations over 30 runs. From Fig.~\ref{fig:ensemble-protein-plot}, one can see how this transfer learning strategy can replace in part the computationally expensive evaluations by generalizing well from just learning on smaller protein sequences. 

\begin{table}[!htb]
      \centering
      \caption{Performance comparison of NAS and state-of-the-art algorithms for real-world protein sequences. The best, worst, mean, and standard deviation (Std) are reported over $30$ runs.
      }
        \resizebox{0.95\textwidth}{!}{%
        \begin{tabular}{l|l|l|l|l|l}
        \midrule
        \makecell {Protein} & {Algorithm} & {Best} & {Worst} & {Mean} & {Std}\\
        \midrule
          & IFABC & $ -2.00830e+00  $&$ -1.53750e+00 $&$ -1.73486e+00 $&$ 9.16579e-02 $\\
          & LSHADE & $ -2.37310e+00  $&$ -1.71710e+00 $&$ -2.10369e+00 $&$ 2.61914e-01 $\\
        $1$BXP & SGDE & $ -2.30300e+00  $&$ -1.26950e+00 $&$ -2.18265e+00 $&$ 2.59131e-01 $\\
         & NAS & $ -2.49023e+00  $&$ -2.07551e+00 $&$ -2.35068e+00 $&$ 1.44632e-01 $\\
        \midrule
          & IFABC & $ -3.01540e+00  $&$ -2.41200e+00 $&$ -2.65486e+00 $&$ 1.42985e-01 $\\
          & LSHADE & $ -4.17940e+00  $&$ -3.88990e-01 $&$ -3.10716e+00 $&$ 8.44159e-01 $\\
        $1$CB$3$ & SGDE & $ -5.09030e-01  $&$ -3.88990e-01 $&$ -4.77019e-01 $&$ 5.30837e-02 $\\
         & NAS & $ -4.19828e+00  $&$ -3.40068e+00 $&$ -3.85454e+00 $&$ 2.80526e-01 $\\
        \midrule
          & IFABC & $ -6.81900e+00  $&$ -5.27940e+00 $&$ -6.07831e+00 $&$ 2.97476e-01 $\\
          & LSHADE & $ -8.34140e+00  $&$ -6.01370e+00 $&$ -7.8585e+00 $&$ 6.67609e-01 $\\
        $1$BXL & SGDE & $ -8.11630e+00  $&$ -5.93430e+00 $&$ -6.65916e+00 $&$ 5.86852e-01 $\\
        & NAS & $ -8.61305e+00  $&$ -7.18074e+00 $&$ -8.19518e+00 $&$ 4.44214e-01 $\\
        \midrule
          & IFABC & $ -4.77300e+00  $&$ -3.36490e+00 $&$ -3.78407e+00 $&$ 2.70870e-01 $\\
          & LSHADE & $ -6.95040e+00  $&$ -2.99190e+00 $&$ -4.78441e+00 $&$ 1.24773e+00 $\\
        $1$EDP & SGDE & $ -6.23270e+00  $&$ -4.53410e-01 $&$ -1.28434e+00 $&$ 1.21523e+00 $\\
        & NAS & $ -6.95038e+00  $&$ -1.23753e+00 $&$ -5.37093e+00 $&$ 1.52153e+00 $\\
        \midrule
          & IFABC & $ -5.82240e+00  $&$ -4.49080e+00 $&$ -5.06992e+00 $&$ 3.44016e-01 $\\
          & LSHADE & $ -7.08230e+00  $&$ -4.15840e+00 $&$ -5.37076e+00 $&$ 6.01159e-01 $\\
        $2$ZNF & SGDE & $ -7.13800e+00  $&$ -4.65980e+00 $&$ -5.42427e+00 $&$ 7.59187e-01 $\\
          & NAS & $ -7.41105e+00  $&$ -3.45382e+00 $&$ -5.93711e+00 $&$ 9.51282e-01 $\\
        \midrule
          & IFABC & $ -7.26590e+00  $&$ -4.35930e+00 $&$ -5.06827e+00 $&$ 5.00148e-01 $\\
          & LSHADE & $ -8.81030e+00  $&$ -2.47920e+00 $&$ -4.80588e+00 $&$ 1.71807e+00 $\\
        $1$EDN & SGDE & $ -8.81030e+00  $&$ -3.54610e+00 $&$ -5.84384e+00 $&$ 1.26685e+00 $\\
        & NAS & $ -8.14153e+00  $&$ -3.76380e+00 $&$ -6.04727e+00 $&$ 7.84394e-01 $\\
        \midrule
          & IFABC & $ -5.98690e+00  $&$ -4.22690e+00 $&$ -4.78016e+00 $&$ 4.24072e-01 $\\
          & LSHADE & $ -7.43270e+00  $&$ -4.39070e+00 $&$ -5.41422e+00 $&$ 1.02956e+00 $\\
        $1$DSQ & SGDE & $ -7.43270e+00  $&$ -3.60300e+00 $&$ -6.18808e+00 $&$ 1.32262e+00 $\\
        & NAS & $ -7.43270e+00  $&$ -5.31426e+00 $&$ -6.85287e+00 $&$ 8.35519e-01 $\\
        \midrule
          & IFABC & $ -1.67640e+01  $&$ -1.42670e+01 $&$ -1.51513e+01 $&$ 6.53529e-01 $\\
          & LSHADE & $ -2.17800e+01  $&$ -1.61490e+01 $&$ -1.91514e+01 $&$ 1.17516e+00 $\\
        $1$SP$7$ & SGDE & $ -2.17790e+01  $&$ -1.64290e+01 $&$ -1.92621e+01 $&$ 1.70078e+00 $\\
         & NAS & $ -2.17526e+01  $&$ -1.69024e+01 $&$ -1.98864e+01 $&$ 1.40123e+00 $\\
        \midrule
          & IFABC & $ -6.04400e+00  $&$ -4.64080e+00 $&$ -5.23771e+00 $&$ 3.16578e-01 $\\
          & LSHADE & $ -6.31350e+00  $&$ -4.15670e+00 $&$ -4.45938e+00 $&$ 6.03639e-01 $\\
        $2$H$3$S & SGDE & $ -6.42170e+00  $&$ -4.15670e+00 $&$ -4.45630e+00 $&$ 6.36553e-01 $\\
         & NAS & $ -7.56085e+00  $&$ -3.42569e+00 $&$ -4.96418e+00 $&$ 1.08122e+00 $\\
        \midrule
          & IFABC & $ -1.01220e+01  $&$ -8.94270e+00 $&$ -9.44181e+00 $&$ 2.62877e-01 $\\
          & LSHADE & $ -1.29300e+01  $&$ -7.46260e+00 $&$ -1.00965e+01 $&$ 1.71148e+00 $\\
        $1$FYG & SGDE & $ -1.38600e+01  $&$ -1.01240e+01 $&$ -1.19298e+01 $&$ 1.08118e+00 $\\
         & NAS & $ -1.40123e+01  $&$ -7.08564e+00 $&$ -1.19396e+01 $&$ 1.51319e+00 $\\
        \midrule

        \end{tabular}}
      \label{table:proteins-res}%
\end{table}%

\vspace{-0.5cm}

\section{Discussion\label{discussion}}

tThe obtained results for CEC $2017$ and PSP show that NAS is comparable in performance to the traditional evolutionary algorithms which call into question the necessity of incorporating the convolutional operations in hyper-heuristic methods. The comprehensive experimental evaluation of transfer learning and ensemble learning allows us to pinpoint the performance gains associated with the NAS evaluation scheme. We conjecture that even better results could be attained if NAS optimizes multiple PSP sequences to learn the joint distribution of all the instances, and then transfers the learned weights to a new protein sequence. Meanwhile, the authors would like to discuss two major takeaways. First, the application of the existing NAS methods without considering the early stopping strategy decreases the accuracy of the reported results. Relatedly, it is difficult to quantify the performance gains without fine-tuning against leading batch size hyperparameter. In a sense, it is actually an abstract way of optimizing the behavior of Adam. Second, unlike  related  approaches, NAS needs more computational time when the computational resources are limited.

\vspace{-0.5cm}

\section{Conclusion and Future Directions \label{conclusion}}

This study revisits the common application of the NAS and reformulates it for optimization tasks. The introduced hyper-heuristic perspective facilitates the process of generating efficient solvers by leveraging a mixture of convolution components from CNNs. We conducted some experiments regarding two aspects: (1) using NAS for tackling standard CEC $2017$ functions and PSP instances, (2) exploring search acceleration possibilities by means of transfer learning and ensemble learning techniques. Empirical results suggest the superiority of NAS for both problems regarding solution accuracy metric. Furthermore, we  conclude  that  NAS can provide an advantage  over  standard  evolutionary algorithms  when transferring is enabled. Accordingly, the trained networks can be employed to deliver useful information on unseen optimization problems without further training. In the future, we will consider NAS for multi-objective and binary search spaces.

\section{Acknowledgments}

The authors would like to acknowledge the High-Performance Computing Center of the University of Strasbourg for supporting this work by providing scientific support and access to computing resources. Part of the computing resources was funded by the Equipex Equip@Meso project (Programme Investissements d'Avenir) and the CPER Alsacalcul/Big Data.

\bibliographystyle{elsarticle-num-names}
{\small
\bibliography{main.bib}}

\end{document}